\documentclass[final,twocolumn]{elsarticle}

\usepackage[letterpaper,margin=2cm]{geometry}
\usepackage{microtype}
\usepackage{geometry}
\usepackage{import}
\usepackage{amssymb}
\usepackage{amsmath}
\usepackage{booktabs}
\usepackage{multirow}
\usepackage{colortbl}
\usepackage{xcolor}
\usepackage{lineno}
\usepackage{array}
\usepackage{rotating}
\usepackage[table]{xcolor}
\usepackage{graphicx}
\usepackage{tikz}

\definecolor{AblationBlue}{HTML}{4C91BC}
\definecolor{AblationOrange}{HTML}{D99053}

\newcommand{\abldeg}[2]{\cellcolor{AblationBlue!#1!white}#2}
\newcommand{\ablimp}[2]{\cellcolor{AblationOrange!#1!white}#2}

\journal{Artificial Intelligence in Agriculture}

\begin{document}

\begin{frontmatter}

\title{MultiAltitudeSR: Multi-Altitude Agricultural UAV Image Super-Resolution via Physical Ground Sampling Distance Conditioning}

\author{Xintong Jiang\fnref{label1}}

\author{Yu Tian\fnref{label1}}

\author{Yixue Liu\fnref{label1}}

\author{Valerio Hoyos-Villegas\fnref{label2}}

\author{Viacheslav Adamchuk\fnref{label1}}

\author[label1]{Shangpeng Sun\corref{cor1}}
\ead{shangpeng.sun@mcgill.ca}

\affiliation[label1]{organization={Department of Bioresource Engineering, McGill University},
            addressline={21111 Lakeshore Road}, 
            city={Ste-Anne-de-Bellevue},
            postcode={H9X 3V9}, 
            state={Quebec},
            country={Canada}}

\affiliation[label2]{organization={Michigan State University},
            addressline={426 Auditorium Road}, 
            city={East Lansing},
            postcode={48824}, 
            state={Michigan},
            country={United States}}

\cortext[cor1]{Corresponding author}

\begin{abstract}
Unmanned aerial vehicles (UAVs) enable high-throughput platforms for crop monitoring but face a fundamental trade-off between flight altitude and image spatial resolution. Many existing super-resolution (SR) methods for UAV imagery rely on oversimplified synthetic degradation models and are restricted to fixed integer upsampling scales, limiting their ability to handle real-world degradations and varying flight altitudes. This study proposes MultiAltitudeSR, a framework that conditions feature modulation and reconstruction on source and target ground sampling distances (GSDs). MultiAltitudeSR trained jointly on real UAV images at altitudes of 10 to 70 m above ground level supports all 21 source-to-target altitude pairs without pair-specific fine-tuning, covering scale factors ranging from 1.18× to 6.03×. The framework can be extended to additional source-target GSD combinations by incorporating corresponding image pairs into training. On the 20-to-10 m pair, the model achieved competitive perceptual performance compared with pair-specific baselines, with LPIPS of 0.2236, DISTS of 0.2431, and NIQE of 8.9020. Experiments using the same architecture demonstrated the value of acquiring real-world multi-altitude UAV image pairs for SR training, with real-pair training achieving better average reconstruction performance across the 21 altitude pairs than training with the tested synthetic degradation pipeline. Downstream zero-shot segmentation case studies illustrated task-dependent benefits of super-resolved imagery for crop monitoring, suggesting potential UAV flight-time savings through image acquisition at higher altitudes.
\end{abstract}


\begin{keyword}
UAV Super Resolution \sep Real-world Image Degradation \sep Multi-altitude UAV Imagery \sep Ground Sampling Distance \sep Plant Phenotyping

\end{keyword}

\end{frontmatter}

\section{Introduction}
Unmanned aerial vehicles (UAVs) have become one of the most promising high-throughput platforms for plant phenotyping and vegetation monitoring at field to regional scales. A major bottleneck of high-throughput UAV monitoring is the trade-off between flight heights and time. While low-altitude high-resolution (HR) UAV images capture fine plant structures and features essential to downstream tasks, their limited ground coverage constrains comprehensive crop monitoring and is thus impractical for breeding population selection. For example, a 2-ha breeding trial takes hours of flight time at 10 m, but only minutes at 70 m. Moreover, UAV acquisition should fall within a narrow window of stable conditions, since image quality depends on weather, solar angle, and various other field factors \citep{gao2021weather}. A multi-hour low-altitude flight cannot yield radiometrically consistent imagery across a single trial. High-altitude UAV low-resolution (LR) images are more practical for real-world production, providing higher throughput but less accurate phenotyping characteristics \citep{MAO2023110698}. Such trade-offs between throughput and accuracy have become a crucial barrier in plant phenotyping and precision agriculture towards the next green revolution \citep{yang2020crop}. 

As a potential solution to this trade-off, deep learning-based super-resolution (SR) in UAV crop imagery has started to draw more attention in recent years \citep{lu2024effective, zhang2025phenosr}. Existing SR methods learn from paired LR and HR images  \citep{wang2020deep, liu2022blind}, and usually comprise three components: a degrader that synthesizes LR inputs from HR references \citep{zhang2021designing, wang2021real}, a generator that restores high-frequency details \citep{wang2018esrgan}, and an up-sampler that projects features to the target resolution \citep{shi2016real}. 

Unlike compression-induced degradations in general computer vision, changes in UAV flight altitude introduce physical degradations that are difficult to characterize and quantify with no explicit function expression \citep{luo2022learning, liu2025real}. Owing to the limited availability of real-world multi-altitude UAV imagery, existing UAV SR studies often rely on synthetic datasets to learn simplified degradation patterns, in which LR images are generated from HR references under simplified degradation assumptions. These assumptions fail to reproduce the altitude-dependent degradation of real UAV imagery, which combines ground sampling distance (GSD) variation, altitude-specific optical blur, and various altitude-related effects such as wind-induced motion blur. Due to this domain gap, models trained on synthetic datasets exhibit degraded performance when transferred to real-world applications. 

Aside from degradation assumptions, SR models in agricultural applications are commonly formulated for predefined integer scales, such as 2$\times$,  3$\times$, and 4$\times$ \citep{lu2024effective, zhang2025phenosr}. These fixed integer scales do not fully reflect the physical resolution characteristics of UAV imagery, which are determined by physical acquisition parameters, including flight altitude and imaging configuration \citep{shi2016unmanned, chakhvashvili2024crop}. Different plant phenotyping tasks require imagery at different physical resolutions \citep{velumani2021estimates, hu2021coupling}. Consequently, predefined scale factors cannot flexibly represent the diverse source-target GSD mappings encountered in practical UAV phenotyping. Furthermore, the same upsampling factor can still correspond to different source and target GSDs, which are indistinguishable when resolution changes are represented solely by scale factors. These characteristics motivate an SR formulation that explicitly accounts for the physical source and target resolutions of UAV imagery.

Collectively, these limitations restrict the translation of conventional SR methods to practical UAV-based plant phenotyping and other image-based agricultural crop-monitoring applications, which require handling both real-world UAV image degradations and flexible resolution mappings. To address these gaps, this study proposes MultiAltitudeSR, a GSD-conditioned multi-scale SR model learning altitude-dependent real-world degradation patterns in UAV crop imagery. By explicitly conditioning the reconstruction process on source and target GSDs, MultiAltitudeSR jointly handles 21 source-target GSD mappings within a single unified model, covering all low-to-high GSD combinations acquired at altitudes from 10 m to 70 m AGL in 10 m increments. Beyond these 21 pairs, MultiAltitudeSR can be readily extended to additional resolution mappings without requiring separate scale-specific models. The main contributions of this work are as follows:

(1) A real-world multi-altitude UAV image dataset constructed across multiple crop types and growth stages, enabling SR to learn practical degradation patterns directly from paired imagery acquired at different flight altitudes.

(2) A GSD-conditioned multi-scale SR framework, MultiAltitudeSR, that jointly handles 21 source-target GSD mappings within a unified model, providing flexibility beyond conventional fixed integer-scale SR. The study interprets the learned SR patterns via advanced visualization techniques and ablation experiments for validation. 

(3) The effects of real-world and synthetic SR datasets are compared under the same SR architecture. Reconstruction of different GSDs and crops is analyzed and interpreted through ablation experiments and visualization.

(4) The potential benefits of SR for downstream crop monitoring are illustrated through zero-shot segmentation case studies spanning different tasks, crop scenes, and source-target GSD mappings.

\section{Methodology}
\subsection{Multi-altitude Crop SR Dataset}
The multi-altitude UAV RGB dataset was acquired using a DJI Mavic 3 Multispectral UAV (SZ DJI Technology Co., Shenzhen, China) with a visible RGB camera (4/3 CMOS, 20MP). RGB imagery was radiometrically corrected together with the multispectral bands using a Sentera Calibrated Reflectance Panel (Sentera LLC, St. Paul, USA) and measurements from the UAV’s onboard sunlight sensor. The dataset includes seven flight altitudes ranging from 10 m to 70 m above ground level (AGL) in 10-meter increments. Orthomosaic maps were reconstructed using Agisoft Metashape (Agisoft LLC., Calgary, Canada), followed by georeferencing, preprocessing and manual validation of image pairing quality. All the flights were conducted with 75\% frontal overlap and 80\% side overlap. 

Despite identical flight and reconstruction settings used throughout the dataset, minor variations in the actual GSDs occurred among reconstructed orthomosaics due to changes in actual acquisition conditions, including local terrain elevation and flight dynamics, as well as in the photogrammetric reconstruction process. Representative GSDs derived from the reconstructed orthomosaics were assigned to each acquisition altitude and used as the conditioning variables. For flight altitudes of 10, 20, 30, 40, 50, 60, and 70 m AGL, the assigned GSDs were 3.3, 5.5, 8.4, 11.3, 14.0, 16.8, and 19.9 mm/pixel, respectively. This discretization enables GSD-aware routing in the proposed framework, with minor inter-flight and reconstruction-induced GSD variations absorbed into the altitude-level representation rather than explicitly calibrated on a per-flight basis. 

\begin{figure*}[!t]
\centering
\includegraphics[width=\textwidth]{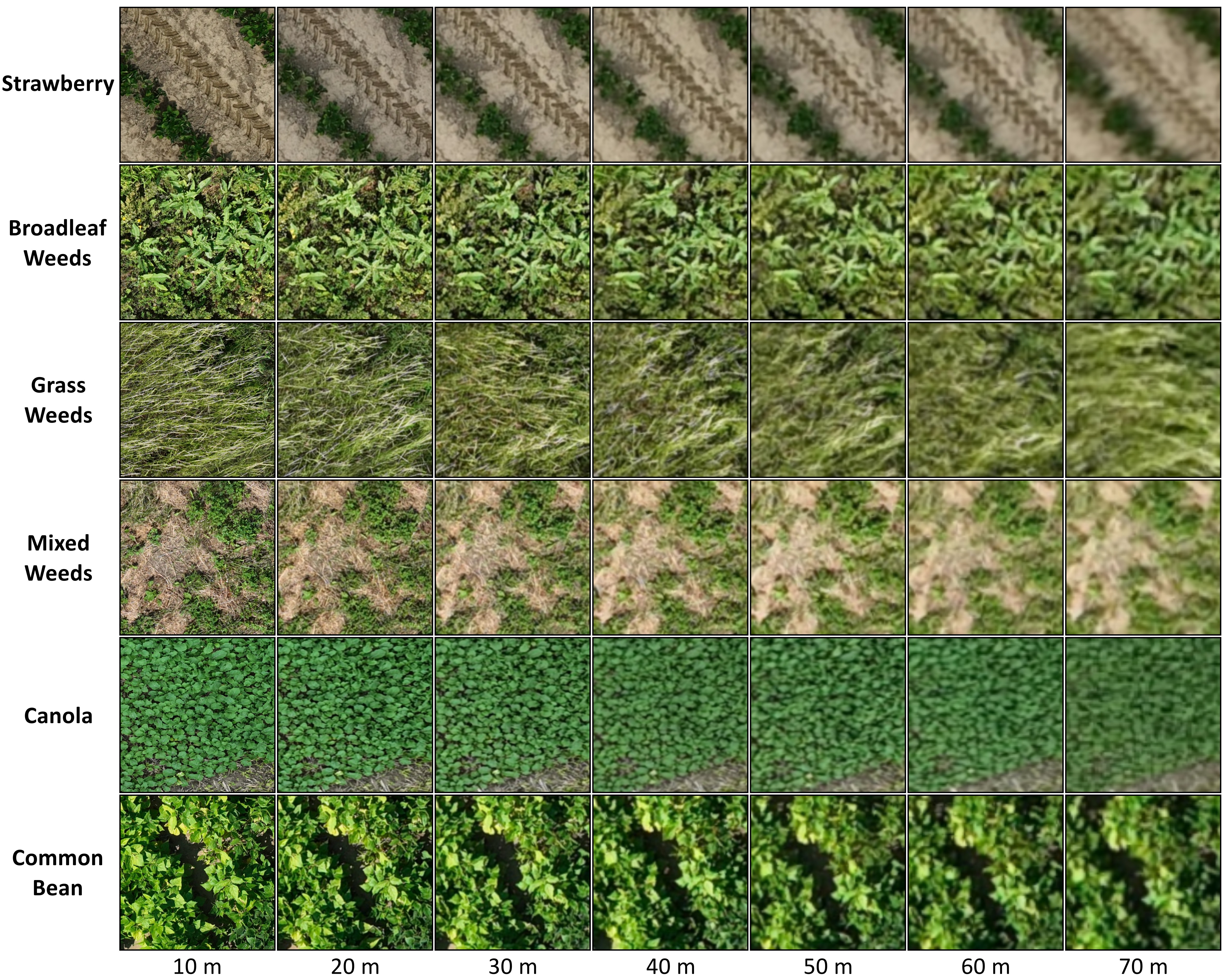} 
\caption{Overview of the multi-altitude UAV dataset that consists of multiple crop and weed species.}
\label{fig: Dataset_overview}
\end{figure*}

\begin{figure*}[!t]
\centering
\includegraphics[width=\textwidth]{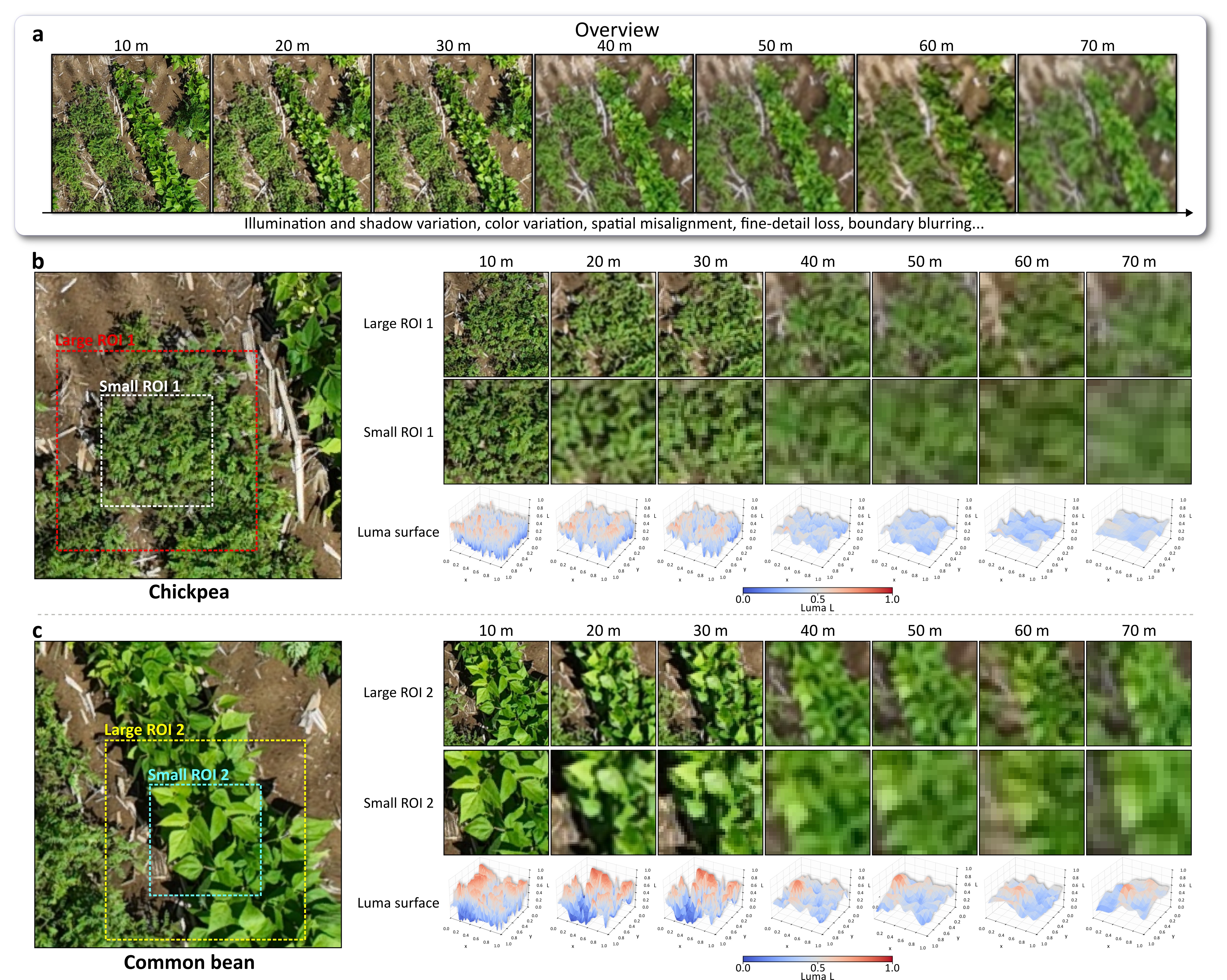} 
\caption{Comparison of image details and luminance surfaces across flight altitudes for common bean and chickpea, two crops with contrasting leaf sizes.}
\label{fig: Chickpea_Common_Bean}
\end{figure*}

Representative samples from the multi-altitude UAV dataset are shown in Fig.~\ref{fig: Dataset_overview}. This dataset covers multiple crops, including common bean, soybean, strawberry, canola, chickpea, and various weeds. To illustrate altitude-dependent changes in fine spatial details and image degradation under field conditions, common bean and chickpea were selected as examples of crops with relatively large and small leaves, respectively. Fig.~\ref{fig: Chickpea_Common_Bean} presents local image details and their corresponding intensity luminance surfaces across flight altitudes. The luma values were computed as $Y' = 0.299R' + 0.587G' + 0.114B'$, where $R'$, $G'$, and $B'$ denote the gamma-encoded RGB channel values \cite{series2011studio}. The dataset was split into training, validation, and test sets at an approximately 8:1:1 ratio, comprising 8,673, 1,050, and 1,029 images, respectively. The three subsets have approximately consistent crop proportions.

\subsection{Synthetic HR-LR Pairs}
Due to practical challenges in acquiring paired UAV imagery of different GSDs, existing SR methods commonly rely on synthetic HR-LR pairs generated through predefined degradation processes \citep{tian2025lightweight, li2024lswinsr, aslahishahri2021spatial}. However, real UAV image degradation arises from complex and interacting factors, including optical effects, platform motion, sensor noise, environmental variations, and subsequent image processing \citep{leem2025practical, sieberth2016automatic, ludwig2020quality}. Consequently, accurately reproducing real-world UAV image degradation through synthetic modeling remains challenging.

\begin{figure*}[!t]
\centering
\includegraphics[width=\textwidth]{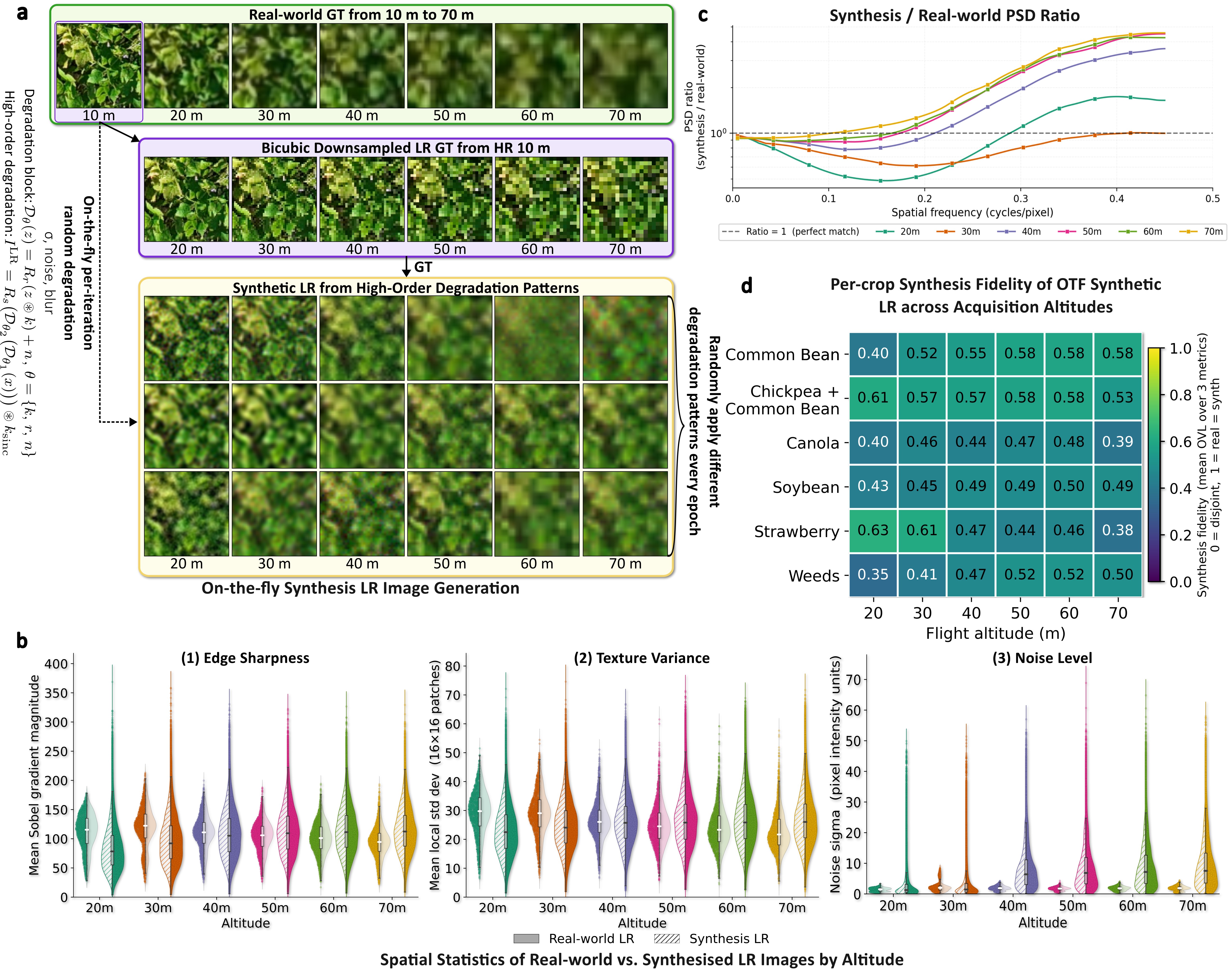} 
\caption{OTF synthetic LR dataset. (a) The OTF synthetic LR generation process, in which only real-world 10 m HR images are used. A random degradation is drawn from the degradation distribution pool every training epoch. Based on each real-world 10 m HR image, a fresh LR input is synthesized by the high-order degradation inside the data loader every epoch during training time. Intermediate GT targets from 20 m to 70 m are bicubic downsamples of the same HR, used as supervision targets during both training and validation. (b) Violin plots of three statistics measuring the difference between the synthetic and real-world datasets. (c) PSD ratio across spatial frequency for each altitude. (d) Fidelity between the real-world and synthetic LR distribution overlaps for different tested crops and altitudes. }
\label{fig: Synthetic_on-the-fly}
\end{figure*}

To validate the necessity of acquiring a real-world multi-altitude dataset, an on-the-fly (OTF) high-order degradation was applied, which is similar to the degradation used in Real-ESRGAN \citep{wang2021real}. A random high-order degradation was generated for every sample at every epoch, as illustrated in Figure~\ref{fig: Synthetic_on-the-fly}(a). To imitate the conditions in which real-world datasets are unavailable, the real-world intermediate altitudes from 20 m to 70 m were also hidden from the model during training and validation. During training and validation, only HR images from the lowest AGL altitude, 10 m, were accessible to the degradation generation process. In addition, no statistic of the real LR domain from 20 m to 70 m was leaked to the model. For these intermediate altitudes, the synthetic model was supervised with a bicubic-downsampled 10 m HR image as pseudo-GT, as in the intermediate pyramid levels of LapSRN \citep{lai2017deep}. All the strength parameters for the random degradation followed Real-ESRGAN's priors. Different from Real-ESRGAN, the JPEG compression step was excluded from our UAV image degradation since Real-ESRGAN's default quality range of 30 to 95 is not valid for our real-world LR images. 

The base-stage degradation block that performs blur, resize and noise is defined as follows:
\begin{equation}
  \mathcal{D}_\theta(i) \;=\; R_r\!\left(i \circledast k\right) + n,
  \qquad \theta = \{k,\,r,\,n\},
  \label{eq:degblock}
\end{equation}
where $i$ denotes the input image, $k$ is a randomly sampled 21 $\times$ 21 blur kernel, $\circledast$ is a 2D convolution with reflect padding, $R_r(\cdot)$ is a random resize operator by factor $r$, and $n$ is additive Gaussian noise. 

The high-order degradation is defined by applying two independently sampled degradation blocks $\mathcal{D}_{\theta}$ in Eq.~\eqref{eq:degblock}:
\begin{equation}
  I^{\mathrm{LR}} \;=\;
  R_s\!\Big(\mathcal{D}_{\theta_2}\!\big(\mathcal{D}_{\theta_1}(x)\big)\Big)
  \circledast k_{\mathrm{sinc}},
  \label{eq:highorder}
\end{equation}
where $I^{\mathrm{LR}}$ denotes the synthesized LR image, $R_s$ is the final resize to the target LR side length $s$, $x$ is the HR source image from 10 m AGL, and $k_{\mathrm{sinc}}$ is the final sinc filter that reintroduces the ringing and overshoot artifacts characteristic of real image pipelines. The two nested blocks in Eq.~\eqref{eq:highorder} give the operator its high-order character. Unlike a single blur-downsample step, the nested composition in Eq. (2) is intended to approximate the compound degradation of real UAV captures.

During the 240 training epochs, the degradation pool yielded around 1.78 million synthetic-LR images to provide various degradation patterns for blind SR to learn from. Figure \ref{fig: Synthetic_on-the-fly}(b) and (c) depict the differences between the OTF synthetic images and the real-world dataset, and Fig. \ref{fig: Synthetic_on-the-fly}(d) shows the per-crop difference across all tested LR altitudes. The power spectral density (PSD) ratio computes the azimuthally averaged 1D PSD ratio of each image \citep{durall2020watch}, measuring how much signal energy the synthetic LR images have compared to their corresponding real-world LR. For every LR image, three per-image statistics were used to evaluate the differences between real and synthetic LR images: (1) edge strength was measured by the mean Sobel gradient magnitude on the grayscale image using 3 $\times$ 3 kernels; (2) local texture by the mean standard deviation of non-overlapping 16 $\times$ 16 grayscale patches; and (3) noise level, the Laplacian noise estimator averaged over the RGB channels \citep{immerkaer1996fast}. For per-crop per-altitude fidelity measurement, the mean overlap coefficient (OVL) was quantified between the real-LR images and the synthetic-LR distribution of these three statistics. For each statistic, both distributions were estimated as area-normalized 64-bin histograms. 

Figure~\ref{fig: Synthetic_on-the-fly} shows that synthetic LR differs significantly from real-world LR, with simulated LR images underestimating mid-frequency energy but overestimating high-frequency energy relative to real-world captures, consistent with the excess noise in the synthetic images. The PSD ratios dropped below 0.5 near 0.2-0.35 cycles/pixel, and per-crop mean-OVL ranged from 0.35 to 0.63 across crops and altitudes. All three tested statistics exhibited different levels of discrepancy across all altitudes, with the noise level of synthetic LR being the farthest from the real LR. Overall, the synthetic and real LR datasets overlapped poorly.

\subsection{Model Overview}
This study proposes MultiAltitudeSR, a GSD-conditioned SR framework designed for multi-altitude UAV imagery. Given an LR input image acquired at a known source altitude and a specified target altitude, the model reconstructs an SR image with a spatial resolution corresponding to the target GSD. Different from fixed-scale SR methods, the model explicitly conditions on the source-to-target GSD pair and adapts its reconstruction pathway accordingly. 

\begin{figure*}[!t]
    \centering
    \includegraphics[width=\textwidth]{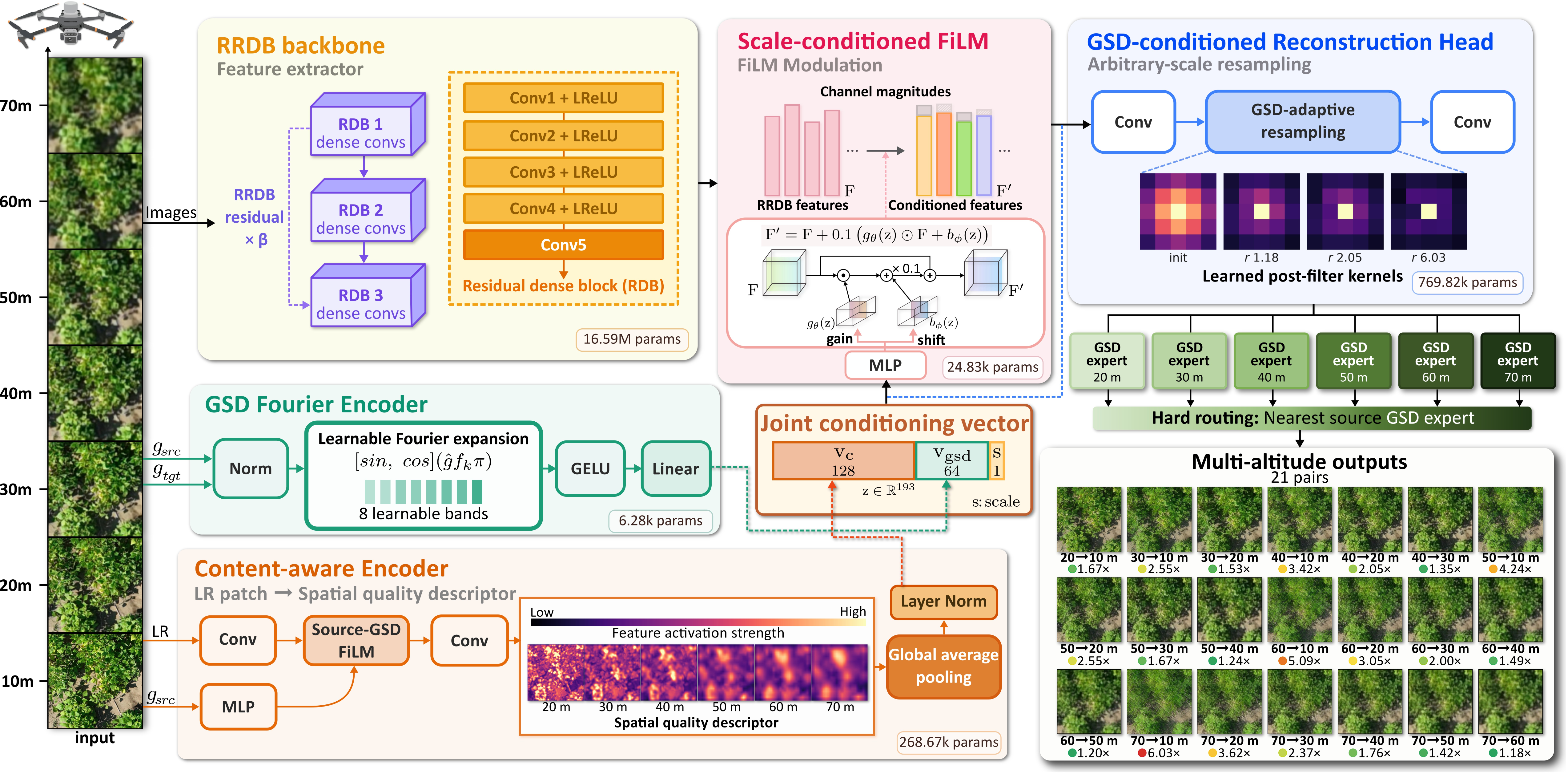}
    \caption{Model overview. UAV imagery of the same region of interest (RoI) is acquired at seven altitudes AGL. An RRDB backbone, initialized with pretrained weights and fine-tuned during training, extracts image features. The content-aware encoder derives a spatial quality descriptor from LR inputs, while the GSD Fourier encoder embeds the source and target GSDs. Together, these encodings condition the modulation of backbone features through scale-conditioned FiLM. The GSD-conditioned reconstruction head applies GSD-conditioned filtering to resample features to the target resolution. Source-GSD-based hard routing then selects one of six output experts to reconstruct the super-resolved image. }
    \label{fig:move_overview}
\end{figure*}

The architecture consists of five components as shown in Fig. \ref{fig:move_overview}: (1) a pretrained Residual-in-Residual Dense Block (RRDB) feature extractor; (2) a GSD Fourier encoder that embeds the source and target GSD values into a scale-aware conditioning vector; (3) a content-aware encoder that infers image-level blur and noise characteristics from the LR input; (4) a Feature-wise Linear Modulation (FiLM) \citep{perez2018film}  that applies an altitude-dependent affine modulation to the backbone features; and (5) a GSD-conditioned reconstruction head that performs upsampling and residual reconstruction under the joint conditioning signal. 

\subsection{RRDB Feature Extractor}
The backbone is an RRDB network initialized from the Real-ESRGANx2plus pretrained model \citep{wang2021real}. It comprises a 3 $\times$ 3 head convolution that maps RGB inputs to 64-channel features, 23 RRDB blocks, and another 3 $\times$ 3 body convolution. Each RRDB contains three residual dense blocks (RDBs) with dense skip connections and a residual scaling factor of 0.2. A global residual connection adds the shallow features to the body convolution output. To adapt the original backbone from a 12-channel pixel-unshuffled input to a 3-channel RGB input, the pretrained input filters are averaged within each corresponding channel group.

Given an LR input $\mathbf{I}_{\mathrm{LR}} \in \mathbb{R}^{B \times 3 \times H \times W}$, the backbone extracts feature maps $\mathbf{F} \in \mathbb{R}^{B \times 64 \times H \times W}$ at the original input resolution. For a given input image, these feature maps are shared across all target GSDs.

\subsection{GSD Fourier Encoder}
Inspired by the Fourier feature encoding of  \citep{tancik2020fourier} that overcomes spectral bias towards low-frequency components of multilayer perceptron (MLP), the GSD Fourier encoder maps the source GSD $g_{s}$ and target GSD $g_{t}$ in mm/pixel to a continuous 64-dimensional embedding vector $\mathbf{v}_{\mathrm{gsd}}$ to encode the magnitude of the scale transition. Raw GSD values are first normalized to $[-1,\,1]$ using the target GSD range. Each normalized scalar is then expanded via Fourier feature encoding with $K = 8$ learnable frequency bands:

\begin{equation}
\begin{aligned}
\hat{g} &= \frac{2(g - g_{\min})}{g_{\max} - g_{\min}}-1,\\
\phi(\hat{g}) &= \left[\sin(\pi f_k \hat{g}),\ \cos(\pi f_k \hat{g})\right]_{k=1}^{K},
\end{aligned}
\label{eq:fourier}
\end{equation}

where  $\hat{g}$ is the normalized GSD, and $f_k \in \mathbb{R}$ are learnable parameters. These $K$ learnable frequency bands allow the encoder to adaptively adjust spectral sensitivities to distinguish GSD-dependent degradation patterns. Concatenating the source and target encoding yields a 32-dimensional feature vector, which is passed through a two-layer MLP to produce $\mathbf{v}_{\mathrm{gsd}} \in \mathbb{R}^{64}$. The learnable frequencies allow the network to discover spectral sensitivities that best discriminate altitude pairs. 

\subsection{Content-aware Encoder}
To complement the GSD-based conditioning derived from the GSD Fourier encoder, the model further encodes content-aware features directly from the low-resolution input. The content-aware encoder produces a compact spatial descriptor of the input image conditioned on the source altitude. It extracts a per-image embedding $\mathbf{v}_{c} \in \mathbb{R}^{128}$ using a lightweight convolutional stack of four layers with LeakyReLU (LReLU) activations and strided downsampling. 

To interpret the extracted features relative to source GSD, a scale-conditioned FiLM layer is applied to condition on the intermediate 128-channel feature map. The same textural pattern is weighted differently depending on the source GSD, reflecting that identical blur may indicate mild degradation at a low altitude but severe degradation at higher altitude. The resulting spatial feature map is globally average-pooled and layer-normalized to yield a compact, stable global descriptor. Unlike the GSD Fourier encoder, which encodes the geometric scale transformation, the content-aware encoder captures image-specific characteristics of the LR input. The two conditioning vectors are concatenated to form a joint conditioning vector:
\begin{equation}
\mathbf{z} = \left[\mathbf{v}_{c};\ \mathbf{v}_{\mathrm{gsd}};\ s\right] \in \mathbb{R}^{193},
\quad s = g_{src}/g_{tgt},
\label{eq:cond}
\end{equation}
which serves as input to both the FiLM modulation layer and the GSD-conditioned reconstruction head.

\subsection{Scale-conditioned FiLM}
The joint conditioning vector $\mathbf{z}$ modulates the RRDB backbone features via Feature-wise Linear Modulation~(FiLM) \citep{perez2018film}. A lightweight MLP predicts per-channel scale and shift offsets, which can be formulated as follows:

\begin{equation}
    \mathbf{F}' = \mathbf{F} + \alpha\left(g_\theta(\mathbf{z}) \odot \mathbf{F} + b_\phi(\mathbf{z})\right),
    \label{eq:film}
\end{equation}
 
\noindent where $\odot$ denotes element-wise multiplication, $g_\theta(\mathbf{z}), b_\phi(\mathbf{z}) \in \mathbb{R}^{C}$ are the per-channel gain and shift predicted by an MLP, and $\alpha = 0.1$ is the residual modulation coefficient. This corresponds to a residual FiLM parameterization ($\gamma = 1 + \alpha g_\theta$,
$\beta = \alpha b_\phi$), with the output layers of $g_\theta$ and $b_\phi$ initialized to zero. This setup ensures that the modulation is identity-preserving at the start of training and is intended to support stable early-stage optimization. The modulated features $\mathbf{F}'$ carry both the structural content from the RRDB backbone and the GSD-specific and content-specific conditioning signal.

\subsection{GSD-conditioned Reconstruction Head} 
\begin{figure*}[!t]
    \centering
    \includegraphics[width=\textwidth]{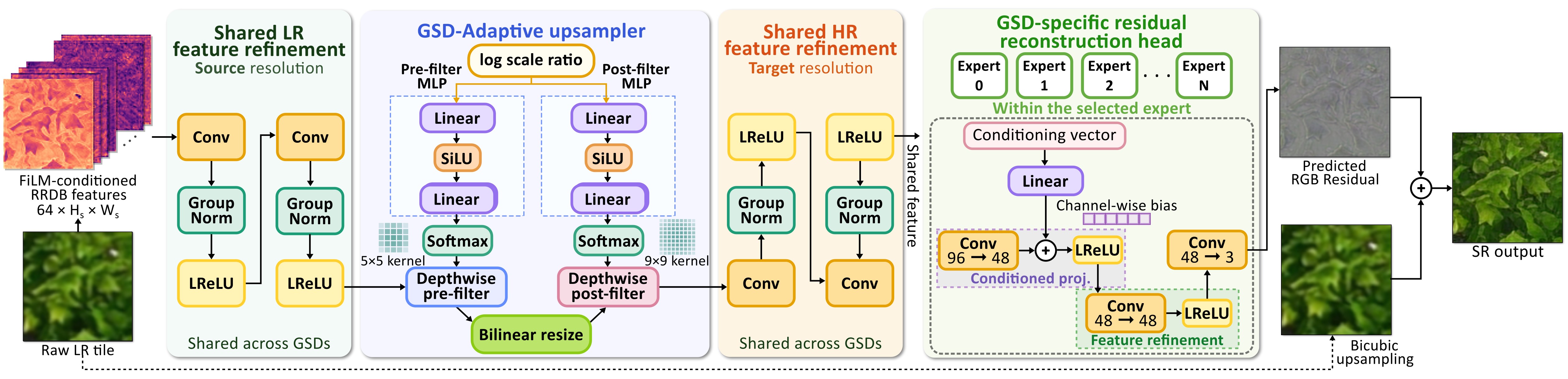}
    \caption{GSD-conditioned adaptive reconstruction head. FiLM-modulated backbone features undergo shared refinement at the source resolution, GSD-adaptive upsampling, and shared refinement at the target resolution. Separate pre-filter and post-filter MLPs map the log source-to-target GSD ratio to eight group-shared 5×5 and 9×9 kernels, respectively, with each kernel normalized by softmax. These depthwise filters operate before and after bilinear resizing. Hard routing selects the expert whose GSD center is nearest to the source GSD. Within the selected expert, a linear projection of the joint conditioning vector $\mathbf{z}$ provides a channel-wise bias, added after the first convolution and before activation. The expert performs conditioned projection, feature refinement, and RGB residual prediction. The predicted RGB residual is added to the bicubic-resized input to produce the super-resolved output.
 }
    \label{fig:reconstruction_head}
\end{figure*}

Fig. \ref{fig:reconstruction_head} illustrates the detailed architecture of the proposed reconstruction head. It operates on FiLM-conditioned RRDB features and performs scale-aware feature refinement, adaptive upsampling, and residual image reconstruction. Given a conditioned feature map $F\in\mathbb{R}^{64\times h\times w}$, the head passes this map through four main components, which are (1) a shared LR Feature Refinement, (2) a GSD-Adaptive Upsampling, (3) a Shared High-Resolution Feature Refinement, and (4) GSD-specific Residual Reconstruction Heads.

\vspace{0.5em}
\noindent\textbf{Shared LR Feature Refinement.}
Before spatial upsampling, the input low-resolution feature map is refined using a lightweight shared convolutional backbone operating at the original source resolution. This is to suppress inter-cell discontinuities and stabilize subsequent interpolation, particularly for large source-to-target scale ratios. This pre-upsampling block consists of repeated convolutional refinement blocks. 

Although 3$\times$3 convolutions are spatially local, their effective receptive field becomes relatively large on small feature maps (e.g., 21 $\times$ 21 pixels), allowing efficient suppression of grid-like structures before upsampling. Stacking multiple 3$\times$3 convolutions further enlarges the effective receptive field while maintaining computational efficiency and strong nonlinear modeling capacity. 

Group Normalization (GN) is applied instead of Batch Normalization due to the varying spatial resolutions applied in multi-GSD training. LReLU with a negative slope of 0.2 is adopted following ESRGAN/Real-ESRGAN-style image restoration architectures due to its stable gradient propagation.

The hidden feature dimension is expanded from 64 to 96 channels to provide sufficient capacity for GSD-conditioned feature modulation and adaptive reconstruction while maintaining practical GPU memory usage. 

\noindent\textbf{GSD-Conditioned Adaptive Upsampling.}
Standard interpolation applies a fixed, scale-agnostic kernel regardless of the source-to-target GSD ratio. At large upsampling ratios, inter-cell discontinuities in low-resolution feature maps are amplified during interpolation, producing periodic grid artifacts that 3$\times$3 convolutions at the target resolution cannot suppress owing to their limited receptive field.
Plain interpolation is therefore replaced by bilinear interpolation bracketed by two depthwise filters whose kernels are predicted from the scale ratio $r = g_{\mathrm{source}} / g_{\mathrm{target}}$:
\begin{equation}
    \mathbf{F}_{\uparrow} = \mathcal{K}_{\mathrm{post}}(r) \circledast \mathrm{Bilinear}\!\left(\mathcal{K}_{\mathrm{pre}}(r) \circledast \mathbf{F},\; H_{t}\times W_{t}\right),
\end{equation}
where $\circledast$ denotes depthwise convolution and $H_t \times W_t$ is the target resolution.
The $5\times5$ pre-filter smooths inter-cell discontinuities at the source resolution before interpolation amplifies them, whereas the $9\times9$ post-filter spans the $\sim$6-pixel cell period at the largest ratio and removes residual periodic artifacts.
Each set of kernels is predicted from $\log r$ by a separate two-layer MLP, as scale acts multiplicatively, and is normalized by a softmax over its taps, so that each kernel is a non-negative averaging kernel that preserves the local feature mean.
The final MLP layers are zero-initialized with biases encoding a Gaussian pre-filter and a near-identity post-filter, so that the module starts as Gaussian-smoothed bilinear upsampling and learns scale-dependent refinements during training.

\noindent\textbf{Shared High-Resolution Feature Refinement.}
Following adaptive upsampling, a second shared convolutional backbone refines the high-resolution feature representation prior to RGB reconstruction. This module provides generic high-resolution refinement, while GSD-specific reconstruction is later handled by the subsequent per-GSD expert heads. The architecture mirrors the pre-upsampling refinement stage and similarly consists of repeated blocks. Operating at the target resolution, this refinement focuses on smoothing artifacts introduced by interpolation and retains correct textures and details for fine-grained spatial detail reconstruction. 

\noindent\textbf{GSD-specific Residual Reconstruction Heads.}
To account for the distinct degradation characteristics of each source altitude, the head employs six expert modules, one per source altitude level (20, 30, 40, 50, 60, and 70~m). Given the shared refined feature map, reconstruction routing is performed according to the source GSD. Under the default hard-routing configuration, the nearest predefined source-GSD expert is selected. Alternative soft and learnable routing strategies are also supported, but not discussed in the current study.  

Each expert reconstruction head consists of three convolutional layers with LReLU activations. The conditioning vector is injected through an additive feature bias in the first layer.  Instead of directly predicting the final RGB image, the head estimates a residual image $\Delta I$ that is added to the bicubic-upsampled source image $I_{bicubic}$. This formulation anchors the prediction to a stable low-frequency baseline, allowing the network to focus on recovering high-frequency texture and GSD-specific structural detail.

\section{Experiments and Results}
\subsection{Training Strategy}
\noindent
The model was trained on an NVIDIA H200 SXM GPU with 141 GB VRAM. The input images from 10 m to 70 m were pre-cropped into patches of 128 $\times$ 128, 77 $\times$ 77, 50 $\times$ 50, 37 $\times$ 37, 30 $\times$ 30, 25 $\times$ 25, and 21 $\times$ 21 pixels, respectively. Each ground location was represented by seven spatially corresponding patches from all altitudes. Training samples were then enumerated by altitude pair, with each sample consisting of one source-target patch pair. Each mini-batch was drawn from a single altitude pair so that only one scale factor was presented per optimization step. To improve robustness to spatial misalignment between altitudes in real-world UAV imagery, the target patch was randomly shifted by up to $\pm 3$ pixels along each axis, with reflection padding used to preserve its size. No shifting was applied during validation or testing.

A batch size of 128 was applied, with an AdamW optimizer for the generator ($\beta_1 = 0.9$, $\beta_2 = 0.999$, $\text{weight decay} = 10^{-5}$). A cosine annealing schedule with warm restarts every 40 epochs controlled the generator's learning rate, which was set to $\eta_G = 10^{-4}$ and annealed to $\eta_{\min} = 10^{-5}$, with each cycle beginning with a three-epoch linear warm-up. A fixed random seed was used for all experiments. 

Training was divided into three stages for stability. In stage 1, epochs 0 to 9, the generator was optimized with a Charbonnier pixel loss and a VGG19-based perceptual loss \citep{simonyan2014very} to establish stable structural reconstruction before introducing adversarial training. During the first five epochs, the pretrained RRDB backbone was kept frozen. The backbone was then unfrozen at epoch 5, and thereafter optimized at the same learning rate as the GSD-conditioned reconstruction head. All 21 altitude pairs were included from epoch 0.

From epoch 10 onward, a U-Net Discriminator with spectral normalization was introduced under the LSGAN objective \citep{mao2017lsgan}. Stage 2 lasted from epochs 10 to 12, in which this discriminator was warmed up to build real-fake separation before the generator received any adversarial gradient. In stage 3 from epochs 13 to 239, the adversarial term was enabled for the generator only on pairs targeting 10 m. The discriminator was optimized with Adam ($\beta_1$ =0.9, $\beta_2=0.99$, no weight decay), and its inputs were reflection-padded to a multiple of 8 to accommodate the smallest target patches at 60 m.

An R1 gradient penalty \citep{mescheder2018training} with $\gamma = 0.05$ was applied lazily to the real branch to suppress overfitting to local texture patterns. Following the lazy-regularization schema applied in StyleGAN2 \citep{karras2020analyzing}, the penalty was evaluated every 16 iterations and scaled by the interval to preserve its effective strength. The discriminator learning rate was set to $0.5 \cdot \eta_G$, tracking the generator schedule, to prevent early discriminator dominance.

\subsection{Loss Function}
Similar to the loss function used in Real-ESRGAN \citep{wang2021real}, the generator was trained with a weighted combination of a pixel fidelity loss, a VGG19-based perceptual loss \citep{simonyan2014very}, and an adversarial loss. Instead of the standard $\ell_1$ pixel loss used in Real-ESRGAN, this study adopted a Charbonnier loss augmented with an explicit channel-wise intensity-matching term \citep{charbonnier1994two, lai2017deep}. 

The pixel fidelity loss is formulated as follows:
\begin{equation}
\begin{aligned}
&\mathcal{L}_{\text{pix}}
= \frac{1}{N}\sum_{i=1}^{N}
  \sqrt{\bigl(\hat{I}_{SR,i}-I_{HR,i}\bigr)^2+\epsilon^2}
\\
&\quad + \frac{1}{BC}\sum_{b,c}
  \bigl|
  \mu_{H,W}(\hat{\mathbf{I}}_{SR})_{b,c}
  - \mu_{H,W}(\mathbf{I}_{HR})_{b,c}
  \bigr|.
\end{aligned}
\end{equation}

where $N = B\times C\times H\times W$ is the total number of elements, $\epsilon = 10^{-3}$ and $\mu_{H,W}$ denotes the spatial mean of channel $c$ of image $b$ computed per image. The Charbonnier term provides smoother gradients than $\ell_1$ near zero residuals. During early training, when the generator head is randomly initialized and adversarial training has not yet stabilized, perceptual gradients may lead to erroneous color shifts. The second term explicitly penalizes per-channel intensity drift to mitigate this color-cast failure mode.

The perceptual loss $\mathcal{L}_\text{perc}$ follows the Real-ESRGAN formulation \citep{wang2021real}, and is computed as the $\ell_1$ distance between VGG19 feature maps extracted before each ReLU at layers $conv_{(1,2)}$, $conv_{(2,2)}$, $conv_{(3,4)}$, $conv_{(4,4)}$, and $conv_{(5,4)}$, with corresponding layer weights of 0.1, 0.1, 1.0, 1.0, 1.0. The combined generator loss function is formulated as follows:

\begin{equation}
\mathcal{L}_G
=
\underbrace{\lambda_\text{pix}\mathcal{L}_\text{pix}}_{\text{Pixel loss}}
+
\underbrace{\lambda_\text{perc}\mathcal{L}_\text{perc}}_{\text{Perceptual loss}}
+
\underbrace{\lambda_\text{adv}\mathcal{L}_\text{adv}}_{\text{Adversarial loss}}.
\label{eq:loss}
\end{equation}

\noindent where $\lambda_\text{adv} = 0$ for epochs 0 to 12 and $\lambda_\text{adv} = 0.05$ after full activation. The adversarial term $\mathcal{L}_\text{adv} = \mathbb{E}[(D(\hat{\mathbf{I}}_{SR}) - 1)^2]$ is the LSGAN generator loss \citep{mao2017lsgan}. Compared with the margin-based hinge loss, the quadratic LSGAN loss provides a smoother adversarial learning signal when discriminator predictions deviate substantially from the target labels. LSGAN loss was found to yield more stable optimization for the real-world multi-altitude UAV image pairs with inherent spatial and temporal misalignment. Loss weights were set to $\lambda_\text{pix} = 1.0$ and $\lambda_\text{perc} = 0.75$ throughout training.

\subsection{Evaluation Metrics}
\noindent
This study evaluates reconstruction quality using seven metrics of both full-reference and no-reference categories. For full-reference evaluation focusing on fidelity, Peak Signal-to-Noise Ratio (PSNR) measures pixel-level fidelity, while Structural Similarity Index Measure (SSIM) assesses structural similarity between the reconstructed SR images and HR ground truth \citep{wang2004image}. Multi-Scale Structural Similarity (MS-SSIM) extends SSIM by evaluating structural similarity across multiple scales, capturing both fine-grained and coarse spatial information \citep{wang2003multiscale}. Mean absolute error (MAE) measures the mean absolute pixel-level deviation between SR and HR. For perceptual quality, learned perceptual image patch similarity (LPIPS) quantifies perceptual similarity by comparing deep feature representations, which provides better alignment with human perceptual judgments than pixel-wise metrics \citep{zhang2018unreasonable}. Deep image structure and texture similarity (DISTS) jointly evaluates structural and textural similarity that is invariant to mild texture resampling while remaining sensitive to structural distortions \citep{ding2020image}. For no-reference evaluation, Natural Image Quality Evaluator (NIQE) assesses image quality based on deviations from natural scene statistics without requiring a ground-truth reference \citep{mittal2012making}. Due to limited patch sizes at 20  m to 70  m, NIQE was computed only for SR outputs targeting 10  m HR. 

\subsection{SoTA Comparison}
\noindent
This study benchmarks the proposed model against four baselines: bicubic interpolation, EDSR \citep{lim2017enhanced}, SwinIR \citep{liang2021swinir}, and Real-ESRGAN \citep{wang2021real}. EDSR, SwinIR, and Real-ESRGAN are fixed-scale architectures designed for fine-detail reconstruction at integer upscaling factors such as 2$\times$ or 4$\times$. Therefore, this comparison was conducted only on 20$\rightarrow$10 m pairs, whose effective scale factor of 1.67$\times$ is the smallest among pairs targeting the 10 m images, and the closest to the 2$\times$ setting supported by all three methods. The 2$\times$ outputs of these models were therefore downsampled to the ground-truth resolution using bicubic interpolation before evaluation. To ensure a fair comparison, these models were fine-tuned on the same training and validation splits until convergence. While these baselines were fine-tuned specifically on the 20$\rightarrow$10 m pairs, the proposed model is a single network trained jointly on all 21 altitude pairs without any pair-specific tuning.

Table \ref{tab:sota_20to10} reports the quantitative comparison between the proposed model and the SoTA baselines on both perceptual and fidelity metrics. Real-ESRGAN achieves the best full-reference perceptual scores in LPIPS and DISTS, while SwinIR attains the best NIQE. The proposed model is the only method ranked within the top two on all three perceptual metrics. Since the proposed model shares the same pretrained RRDB backbone as Real-ESRGAN, the comparison between them partly reflects the cost of multi-altitude generalization. As a single model covering all 21 altitude pairs, the proposed model falls only slightly behind Real-ESRGAN, with an LPIPS of 0.2236 versus 0.2144 and a DISTS of 0.2431 versus 0.2365, corresponding to increases of only 4.3\% and 2.8\%, respectively. In contrast, it achieves a better NIQE of 8.90 versus 9.43, a higher PSNR of 16.04 versus 15.78 dB, and a higher MS-SSIM of 0.6497 versus 0.6473. These results indicate that supporting all altitude pairs through GSD conditioning compromises perceptual quality only marginally. Overall, the proposed model achieves competitive and balanced performance across the perceptual metrics.

For all models, fidelity metrics are naturally modest, as the dataset contains pixel-level misalignment arising from real-world spatial and temporal variation between captured altitudes. This misalignment particularly penalizes structure-sensitive metrics such as SSIM, under which bicubic interpolation outperforms all generative models and the proposed model scores lowest. Although EDSR leads on all pixel-level fidelity metrics, its L1-supervised training objective biases it toward mean regression at the cost of perceptual sharpness, as reflected in its LPIPS and DISTS, which are worse even than those of bicubic interpolation.

\begin{table*}[!tp]
\centering
\caption{Quantitative comparison with state-of-the-art methods on the 20$\rightarrow$10 m altitude pair. Bold values indicate the best performance in each column.}
\label{tab:sota_20to10}
\setlength{\tabcolsep}{8pt}
\renewcommand{\arraystretch}{1}
\begin{tabular}{l ccc ccc}
\toprule
 & \multicolumn{3}{c}{Perceptual} & \multicolumn{3}{c}{Fidelity} \\
\cmidrule(lr){2-4}\cmidrule(lr){5-7}
Method & LPIPS$\downarrow$ & DISTS$\downarrow$ & NIQE$\downarrow$
       & PSNR (dB)$\uparrow$ & SSIM$\uparrow$ & MS-SSIM$\uparrow$ \\
\midrule
Bicubic     & 0.3961 & 0.2706 & 14.5626 & 15.84 & 0.2945 & 0.6499 \\
EDSR        & 0.5393 & 0.3406 & 11.0497 & \textbf{16.83} & \textbf{0.3020} & \textbf{0.6631} \\
SwinIR      & 0.2265 & 0.2503 & \textbf{7.7757} & 16.16 & 0.2747 & 0.6421 \\
Real-ESRGAN & \textbf{0.2144} & \textbf{0.2365} & 9.4331 & 15.78 & 0.2763 & 0.6473 \\
\midrule
Ours        & 0.2236 & 0.2431 & 8.9020 & 16.04 & 0.2684 & 0.6497 \\
\bottomrule
\end{tabular}

\vspace{1mm}
\begin{minipage}{\textwidth}
\centering
\footnotesize
$\uparrow$ indicates that higher is better, and $\downarrow$ indicates that lower is better.
\end{minipage}
\end{table*}

\begin{figure*}[!t]
\centering
\includegraphics[width=\textwidth]{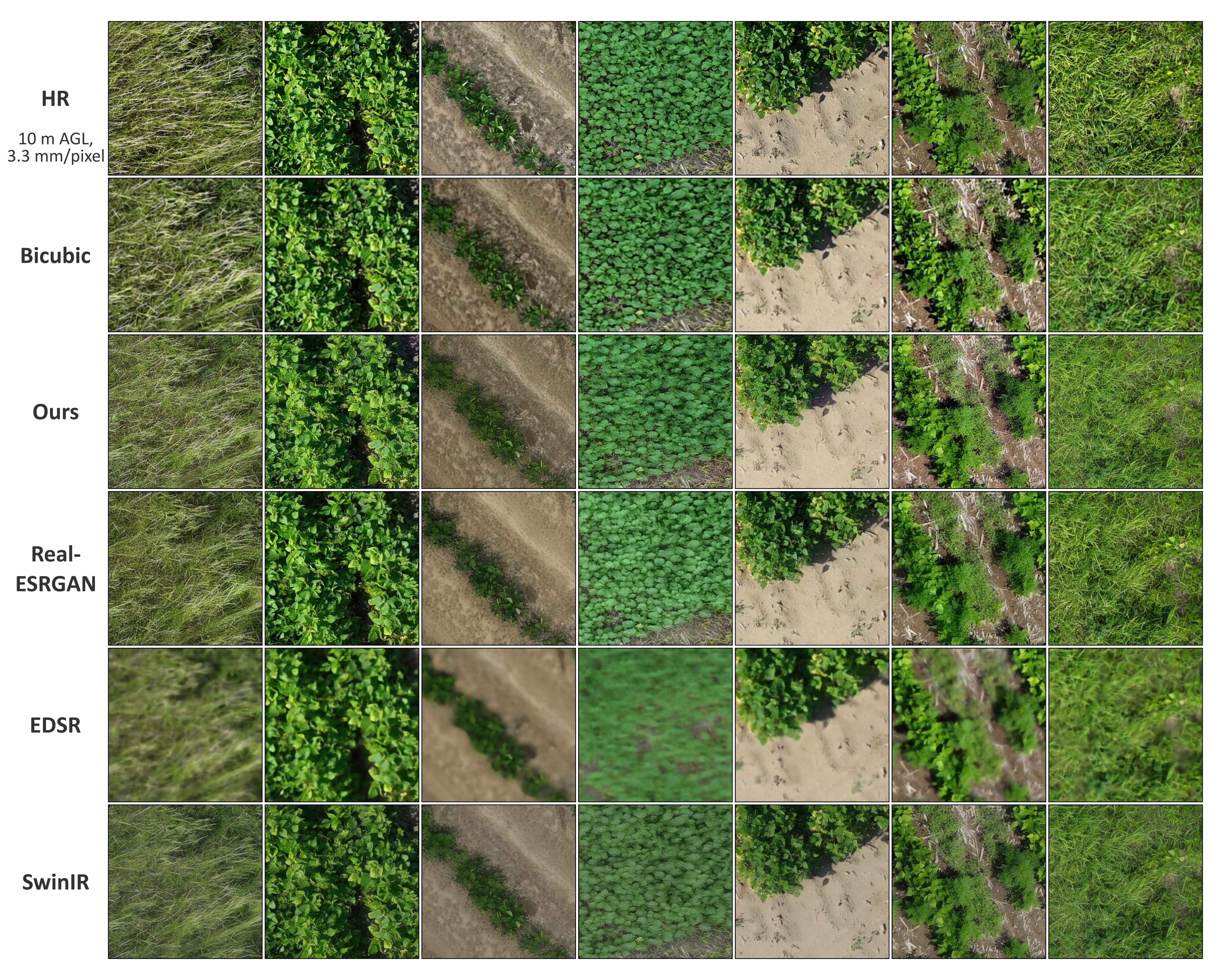} 
\caption{Qualitative comparison of 20$\rightarrow$10 m super-resolution results across seven representative scenes. Each column shows the same tile as the HR reference and as reconstructed by bicubic interpolation and the different super-resolution models.}
\label{fig: SoTA}
\end{figure*}

\begin{figure*}[!t]
\centering
\includegraphics[width=\textwidth]{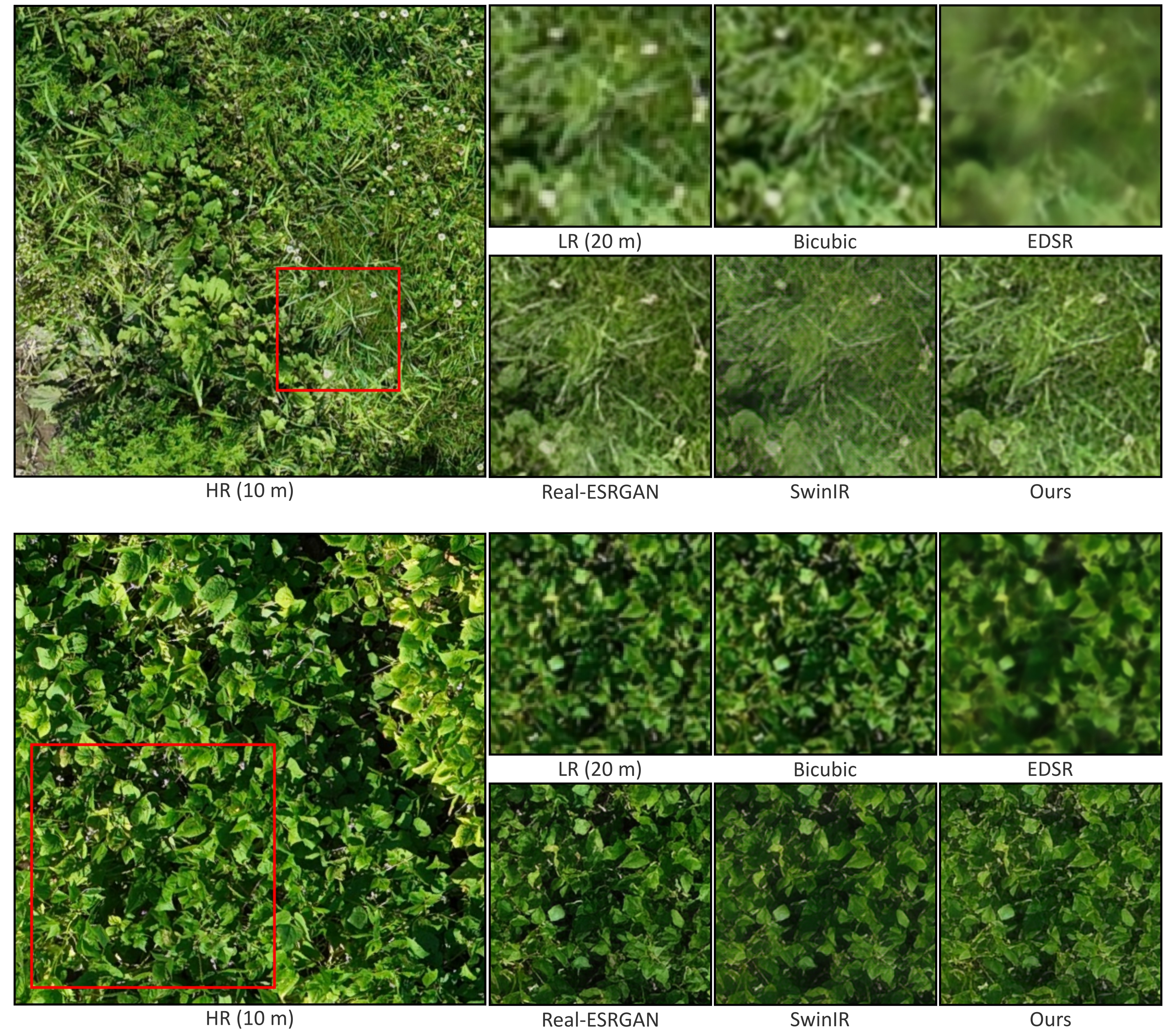} 
\caption{Zoomed-in comparison of 20$\rightarrow$10 m super-resolution results for two representative scenes. The left panel shows the HR reference captured at 10 m AGL, with the red box marking the magnified region. The right panels show the corresponding region from the 20 m LR input and the outputs of bicubic interpolation, EDSR, Real-ESRGAN, SwinIR, and the proposed model.}
\label{fig: zoom-in_SoTA}
\end{figure*}

Figs. \ref{fig: SoTA} and \ref{fig: zoom-in_SoTA} present qualitative comparisons of the 20$\rightarrow$10 m reconstructions. Fig. \ref{fig: SoTA} provides an overview across seven representative scenes covering different plant types and canopy conditions, with each column showing the same tile reconstructed by each method. Fig. \ref{fig: zoom-in_SoTA} complements this overview with zoom-in views of two complicated scenes. In both figures, EDSR results appear over-smoothed and perceptually poor, lacking the fine structural detail recovered by the perceptually oriented approaches. In the zoomed regions, EDSR retains even less visible structure than bicubic interpolation. SwinIR, Real-ESRGAN, and the proposed model all restore sharper leaf boundaries and better canopy texture than EDSR and bicubic interpolation. These qualitative observations are consistent with the quantitative results in Table \ref{tab:sota_20to10}.

\subsection{On-the-fly Synthetic Dataset}
Table \ref{tab:synth_model_syn_vs_real} compares the performance of the OTF model on synthetic and real LR test sets. The synthetic test set was generated by applying the same OTF degradation pipeline to the 147 test-split 10 m HR tiles. Following Real-ESRGAN \citep{wang2021real}, a fresh random degradation was sampled for each training sample, defined by an HR tile and an altitude pair, at every epoch. Each sample was thus exposed to 240 independently sampled degradations over the 240 training epochs. To reduce the sensitivity of the synthetic evaluation to individual degradation draws, 240 independent degradations were similarly sampled for each test image in each altitude, yielding 740,880 synthetic LR inputs from 147 test images $\times$ 21 altitude pairs $\times$ 240 draws. On this synthetic test set, evaluated against the aligned bicubic pseudo-GT, the model achieves markedly higher full-reference scores than on the real LR test set.

\definecolor{synclr}{RGB}{50, 76, 199}
\begin{table*}[t]
\centering
\small
\setlength{\tabcolsep}{3pt}
\caption{SR evaluation of the \emph{synthesis-trained} model, grouped by target altitude. Each cell reports the score on the \textcolor{synclr}{synthetic}\,/\,real test set; MAE is computed on intensities normalized to $[0,1]$, and Mean is computed over all 21 pairs. The lower panel reports the no-reference NIQE, which was computed only for the 10\,m target and is therefore averaged over 6 pairs.}
\label{tab:synth_model_syn_vs_real}
\begin{tabular*}{\linewidth}{@{\extracolsep{\fill}}ll*{6}{c}@{}}
\toprule
\multirow{2}{*}{\textbf{Target}} & \multirow{2}{*}{\textbf{Source}} & \multicolumn{6}{c}{\textcolor{synclr}{\textbf{Synthetic}}\,/\,\textbf{Real}} \\
\cmidrule(lr){3-8}
 & & \textbf{PSNR} (dB)\,$\uparrow$ & \textbf{SSIM}\,$\uparrow$ & \textbf{MS-SSIM}\,$\uparrow$ & \textbf{MAE}\,$\downarrow$ & \textbf{LPIPS}\,$\downarrow$ & \textbf{DISTS}\,$\downarrow$ \\
\midrule
\multirow{6}{*}{10\,m} & 20\,m & \textcolor{synclr}{20.01}\,/\,15.49 & \textcolor{synclr}{0.5944}\,/\,0.2755 & \textcolor{synclr}{0.8803}\,/\,0.6460 & \textcolor{synclr}{0.0811}\,/\,0.1329 & \textcolor{synclr}{0.1826}\,/\,0.2257 & \textcolor{synclr}{0.2035}\,/\,0.2277 \\
 & 30\,m & \textcolor{synclr}{19.17}\,/\,14.99 & \textcolor{synclr}{0.5197}\,/\,0.2159 & \textcolor{synclr}{0.8509}\,/\,0.5957 & \textcolor{synclr}{0.0890}\,/\,0.1412 & \textcolor{synclr}{0.2035}\,/\,0.2703 & \textcolor{synclr}{0.2183}\,/\,0.2597 \\
 & 40\,m & \textcolor{synclr}{18.79}\,/\,14.95 & \textcolor{synclr}{0.4893}\,/\,0.1962 & \textcolor{synclr}{0.8340}\,/\,0.5738 & \textcolor{synclr}{0.0927}\,/\,0.1419 & \textcolor{synclr}{0.2118}\,/\,0.3059 & \textcolor{synclr}{0.2249}\,/\,0.2878 \\
 & 50\,m & \textcolor{synclr}{18.42}\,/\,14.98 & \textcolor{synclr}{0.4458}\,/\,0.1808 & \textcolor{synclr}{0.8133}\,/\,0.5530 & \textcolor{synclr}{0.0967}\,/\,0.1424 & \textcolor{synclr}{0.2338}\,/\,0.3318 & \textcolor{synclr}{0.2315}\,/\,0.3001 \\
 & 60\,m & \textcolor{synclr}{17.96}\,/\,14.81 & \textcolor{synclr}{0.3986}\,/\,0.1752 & \textcolor{synclr}{0.7881}\,/\,0.5361 & \textcolor{synclr}{0.1017}\,/\,0.1467 & \textcolor{synclr}{0.2568}\,/\,0.3517 & \textcolor{synclr}{0.2424}\,/\,0.3149 \\
 & 70\,m & \textcolor{synclr}{17.59}\,/\,14.79 & \textcolor{synclr}{0.3568}\,/\,0.1688 & \textcolor{synclr}{0.7624}\,/\,0.5256 & \textcolor{synclr}{0.1062}\,/\,0.1467 & \textcolor{synclr}{0.3048}\,/\,0.3799 & \textcolor{synclr}{0.2581}\,/\,0.3303 \\
\midrule
\multirow{5}{*}{20\,m} & 30\,m & \textcolor{synclr}{19.02}\,/\,18.11 & \textcolor{synclr}{0.6145}\,/\,0.4783 & \textcolor{synclr}{0.9043}\,/\,0.8037 & \textcolor{synclr}{0.0904}\,/\,0.0982 & \textcolor{synclr}{0.1388}\,/\,0.2327 & \textcolor{synclr}{0.2054}\,/\,0.2504 \\
 & 40\,m & \textcolor{synclr}{18.68}\,/\,17.04 & \textcolor{synclr}{0.5879}\,/\,0.3649 & \textcolor{synclr}{0.8920}\,/\,0.7321 & \textcolor{synclr}{0.0937}\,/\,0.1125 & \textcolor{synclr}{0.1440}\,/\,0.2684 & \textcolor{synclr}{0.2069}\,/\,0.2850 \\
 & 50\,m & \textcolor{synclr}{18.27}\,/\,16.58 & \textcolor{synclr}{0.5435}\,/\,0.2960 & \textcolor{synclr}{0.8754}\,/\,0.6844 & \textcolor{synclr}{0.0983}\,/\,0.1191 & \textcolor{synclr}{0.1621}\,/\,0.3482 & \textcolor{synclr}{0.2195}\,/\,0.3051 \\
 & 60\,m & \textcolor{synclr}{17.80}\,/\,16.37 & \textcolor{synclr}{0.4986}\,/\,0.2886 & \textcolor{synclr}{0.8567}\,/\,0.6688 & \textcolor{synclr}{0.1036}\,/\,0.1231 & \textcolor{synclr}{0.1851}\,/\,0.4009 & \textcolor{synclr}{0.2299}\,/\,0.3168 \\
 & 70\,m & \textcolor{synclr}{17.36}\,/\,16.21 & \textcolor{synclr}{0.4529}\,/\,0.2642 & \textcolor{synclr}{0.8352}\,/\,0.6513 & \textcolor{synclr}{0.1090}\,/\,0.1251 & \textcolor{synclr}{0.1890}\,/\,0.3615 & \textcolor{synclr}{0.2424}\,/\,0.3290 \\
\midrule
\multirow{4}{*}{30\,m} & 40\,m & \textcolor{synclr}{18.55}\,/\,17.59 & \textcolor{synclr}{0.6667}\,/\,0.4705 & \textcolor{synclr}{0.9195}\,/\,0.7908 & \textcolor{synclr}{0.0952}\,/\,0.1066 & \textcolor{synclr}{0.0841}\,/\,0.2337 & \textcolor{synclr}{0.1816}\,/\,0.2832 \\
 & 50\,m & \textcolor{synclr}{18.08}\,/\,17.10 & \textcolor{synclr}{0.6259}\,/\,0.4035 & \textcolor{synclr}{0.9062}\,/\,0.7508 & \textcolor{synclr}{0.1004}\,/\,0.1130 & \textcolor{synclr}{0.0976}\,/\,0.2586 & \textcolor{synclr}{0.1900}\,/\,0.3041 \\
 & 60\,m & \textcolor{synclr}{17.77}\,/\,16.96 & \textcolor{synclr}{0.5901}\,/\,0.3920 & \textcolor{synclr}{0.8941}\,/\,0.7311 & \textcolor{synclr}{0.1040}\,/\,0.1155 & \textcolor{synclr}{0.1076}\,/\,0.2893 & \textcolor{synclr}{0.2027}\,/\,0.3114 \\
 & 70\,m & \textcolor{synclr}{17.24}\,/\,16.73 & \textcolor{synclr}{0.5437}\,/\,0.3601 & \textcolor{synclr}{0.8732}\,/\,0.7158 & \textcolor{synclr}{0.1105}\,/\,0.1182 & \textcolor{synclr}{0.1204}\,/\,0.3345 & \textcolor{synclr}{0.2101}\,/\,0.3245 \\
\midrule
\multirow{3}{*}{40\,m} & 50\,m & \textcolor{synclr}{18.05}\,/\,17.65 & \textcolor{synclr}{0.6860}\,/\,0.4992 & \textcolor{synclr}{0.9383}\,/\,0.8337 & \textcolor{synclr}{0.1009}\,/\,0.1046 & \textcolor{synclr}{0.0912}\,/\,0.2928 & \textcolor{synclr}{0.1802}\,/\,0.3089 \\
 & 60\,m & \textcolor{synclr}{17.64}\,/\,17.57 & \textcolor{synclr}{0.6512}\,/\,0.4896 & \textcolor{synclr}{0.9280}\,/\,0.8249 & \textcolor{synclr}{0.1057}\,/\,0.1059 & \textcolor{synclr}{0.1099}\,/\,0.2915 & \textcolor{synclr}{0.1916}\,/\,0.3199 \\
 & 70\,m & \textcolor{synclr}{17.11}\,/\,17.25 & \textcolor{synclr}{0.6097}\,/\,0.4537 & \textcolor{synclr}{0.9154}\,/\,0.8079 & \textcolor{synclr}{0.1122}\,/\,0.1094 & \textcolor{synclr}{0.1253}\,/\,0.3037 & \textcolor{synclr}{0.1992}\,/\,0.3353 \\
\midrule
\multirow{2}{*}{50\,m} & 60\,m & \textcolor{synclr}{17.69}\,/\,17.65 & \textcolor{synclr}{0.6979}\,/\,0.5328 & \textcolor{synclr}{0.9054}\,/\,0.7845 & \textcolor{synclr}{0.1051}\,/\,0.1055 & \textcolor{synclr}{0.1028}\,/\,0.3346 & \textcolor{synclr}{0.1619}\,/\,0.2967 \\
 & 70\,m & \textcolor{synclr}{17.14}\,/\,17.48 & \textcolor{synclr}{0.6614}\,/\,0.5199 & \textcolor{synclr}{0.8889}\,/\,0.7749 & \textcolor{synclr}{0.1118}\,/\,0.1070 & \textcolor{synclr}{0.1170}\,/\,0.3535 & \textcolor{synclr}{0.1689}\,/\,0.3129 \\
\midrule
60\,m & 70\,m & \textcolor{synclr}{17.20}\,/\,18.27 & \textcolor{synclr}{0.7066}\,/\,0.6360 & \textcolor{synclr}{0.9193}\,/\,0.8710 & \textcolor{synclr}{0.1112}\,/\,0.0972 & \textcolor{synclr}{0.1125}\,/\,0.3875 & \textcolor{synclr}{0.1666}\,/\,0.3023 \\
\midrule
\multicolumn{2}{l}{\textbf{Mean}} & \textcolor{synclr}{18.07}\,/\,16.60 & \textcolor{synclr}{0.5686}\,/\,0.3648 & \textcolor{synclr}{0.8753}\,/\,0.7074 & \textcolor{synclr}{0.1009}\,/\,0.1197 & \textcolor{synclr}{0.1562}\,/\,0.3122 & \textcolor{synclr}{0.2065}\,/\,0.3003 \\
\bottomrule
\end{tabular*}
\vspace{0.9em}
\begin{tabular*}{\linewidth}{@{\extracolsep{\fill}}l*{7}{c}@{}}
\toprule
\textbf{NIQE}\,$\downarrow$ & \multicolumn{6}{c}{\textbf{Source}} & \multirow{2}{*}{\textbf{Mean}} \\
\cmidrule(lr){2-7}
(\textbf{Target}\,$=$\,10\,m) & \textbf{20\,m} & \textbf{30\,m} & \textbf{40\,m} & \textbf{50\,m} & \textbf{60\,m} & \textbf{70\,m} & \\
\midrule
\textcolor{synclr}{\textbf{Synthetic}} & \textcolor{synclr}{10.2694} & \textcolor{synclr}{9.2968} & \textcolor{synclr}{9.9002} & \textcolor{synclr}{10.2778} & \textcolor{synclr}{9.4026} & \textcolor{synclr}{8.3507} & \textcolor{synclr}{9.5829} \\
\textbf{Real} & 11.2666 & 9.7842 & 10.2622 & 10.1173 & 9.1686 & 8.3803 & 9.8299 \\
\bottomrule
\end{tabular*}

\vspace{1mm}
\begin{minipage}{\textwidth}
\centering
\footnotesize
$\uparrow$ indicates that higher is better, and $\downarrow$ indicates that lower is better.
\end{minipage}

\end{table*}

As shown in Table \ref{tab:synth_model_syn_vs_real}, this divergence is reflected in the performance differences between the models trained and validated on these two datasets. The synthesis-trained model degrades substantially on the real test set across most metrics, particularly in PSNR. Synthetic LR slightly improves LPIPS for most 10 m tasks, suggesting a stronger synthetic perceptual reconstruction, but consistently sacrifices pixel fidelity and generalization to real UAV observations.

\begin{figure*}[!t]
\centering
\includegraphics[width=\textwidth]{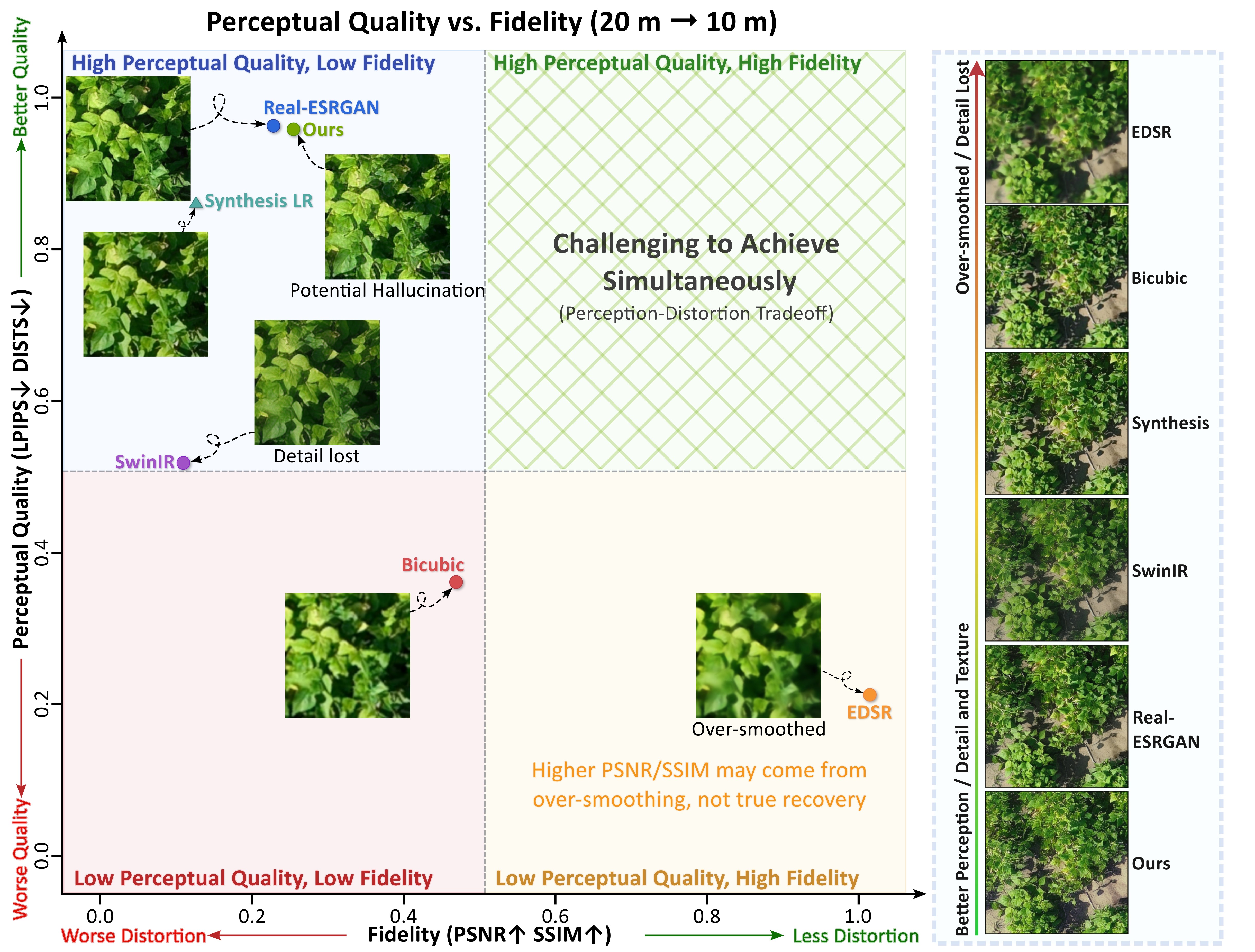} 
\caption{Perception-fidelity comparison on the 20$\rightarrow$10\,m pair. All tested SoTA models, the proposed model trained on real-world pairs (Ours), and the same architecture trained on the synthetic LR dataset (Synthesis LR) are placed on a perceptual-quality (LPIPS$\downarrow$, DISTS$\downarrow$) versus fidelity (PSNR$\uparrow$, SSIM$\uparrow$) plane, with both axes normalized to $[0,1]$ so that larger values indicate better quality. Representative SR results are shown beside each method and in the right column. EDSR achieves the highest PSNR and SSIM, and bicubic interpolation also scores relatively high on SSIM, yet both produce over-smoothed outputs, illustrating that higher fidelity scores do not necessarily correspond to better perceptual reconstruction.}
\label{fig:perceptual}
\end{figure*}

Fig. \ref{fig:perceptual} further illustrates the perceptual-fidelity trade-off among the evaluated models. Although EDSR and bicubic interpolation achieve relatively high PSNR and SSIM, their reconstructed images are visibly smoother and contain few high-frequency details, indicating that high fidelity scores alone do not necessarily correspond to better perceptual reconstruction. In contrast, the proposed model trained on real-world LR observations achieves consistently higher PSNR than its synthesis-trained counterpart while preserving fine spatial details. This suggests that its fidelity improvement does not arise simply from excessive smoothing, but rather from better adaptation to the degradation characteristics of real UAV imagery.

\subsection{Multi-altitude Reconstruction}
\noindent
Table~\ref{tab:real_model_real} presents quantitative evaluation results across all 21 source-to-target GSD pairs, grouped by target resolution. A consistent trend observed across all groups is that performance degrades largely monotonically as the source altitude increases within each target group, reflecting the increasing difficulty of larger super-resolution factors. Within the 10 m target group, the most challenging pair is from 70$\rightarrow$10 m. 

Despite the degradation, PSNR values remain relatively stable across the 10 m group ($\Delta\text{PSNR} < 0.5$ dB), whereas perceptual metrics such as LPIPS and DISTS exhibit pronounced sensitivity to the SR factor, increasing by up to 93\% and 35\%, respectively. At coarser target resolutions, performance improves substantially. The 70$\rightarrow$60 m pair attains the highest PSNR of 20.81 dB and SSIM of 0.7310 in the entire evaluation, which is consistent with the reduced reconstruction difficulty at a 1.18$\times$ spatial ratio. Notably, within the 40 m target group, the 50 m and 60 m source pairs yield nearly identical PSNR values, 19.54 dB and 19.45 dB, suggesting a saturation effect at small upsampling ratios with high target altitudes. In this scenario, a slight decrease in source altitude has a diminishing effect on reconstruction quality.

\begin{figure*}[!t]
\centering
\includegraphics[width=\textwidth]{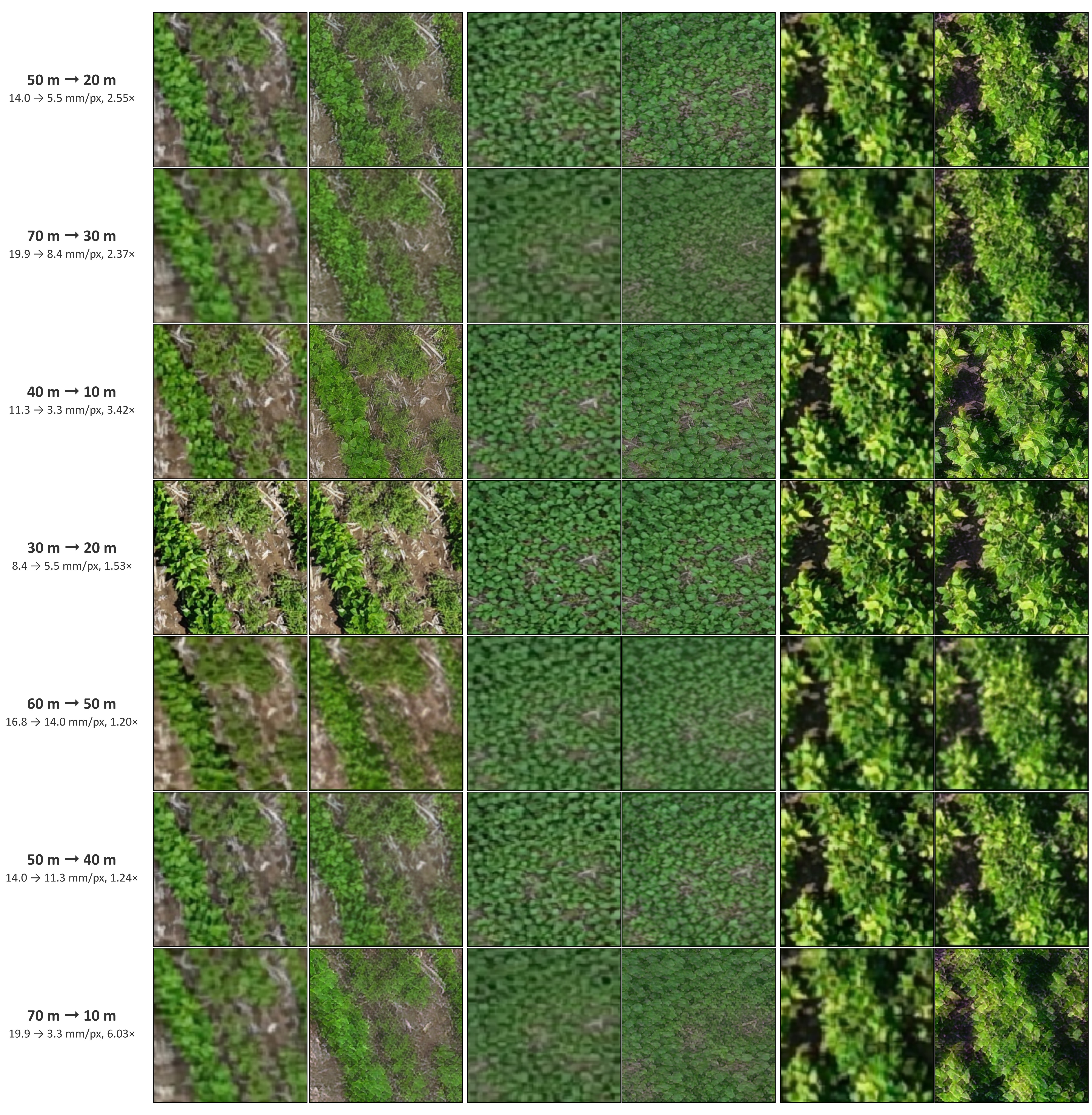} 
\caption{Qualitative super-resolution results for seven representative source-to-target altitude pairs, spanning SR scale factors from $1.20\times$ (60$\rightarrow$50 m) to $6.03\times$ (70$\rightarrow$10 m). Each row shows three representative crops (left: nearest neighbor interpolation of low-resolution images; right: our SR outputs) drawn from distinct scene regions to capture varied canopy density and ground cover composition. GSD values (mm/pixel) for each pair are indicated on the left.}
\label{fig: multi-altitude}
\end{figure*}

NIQE scores are reported only for the 10 m target group, for which patch sizes are sufficiently large for reliable estimation. The scores range from 8.8033 to 9.3105 and show a mild increasing trend with source altitude. Although the SR outputs exhibit visible texture and apparent detail in Fig. \ref{fig: multi-altitude}, these features do not necessarily represent faithfully recovered image structures. Higher NIQE scores at larger upsampling ratios indicate greater deviation from natural image statistics and may reflect reconstruction artifacts, blur, or a combination of both. Therefore, NIQE alone cannot determine whether the observed increase is primarily attributable to artifacts or reduced sharpness.

Fig. \ref{fig: multi-altitude} presents qualitative results across seven representative altitude pairs from different scenes. Reconstruction quality varies with both source and target GSDs, with more pronounced artifacts at finer target GSDs, particularly when the gap between source and target altitudes is large. Fine textures on leaf and soil surfaces may be poorly resolved in the LR input, leaving the model to infer their appearance from patterns learned during training. The resulting textures may appear plausible but fail to represent the actual surface details.

This pattern is most evident in the 10 m AGL target group. The 40$\rightarrow$10 m pair with 3.42$\times$ GSD scale produces well-defined canopy structures with clearly delineated leaf boundaries, whereas the 70$\rightarrow$10 m pair (6.03$\times$) exhibits substantially more severe artifacts. With a source GSD of 19.9 mm/pixel at 70 m AGL, each LR pixel corresponds to approximately 36 output pixels at the target GSD of 3.3 mm/pixel, leaving considerable ambiguity in the spatial details reconstructed within that area. The challenge also relates to the types of detail visible at the target resolution. The 10 m AGL imagery captures within-leaf features, including venation patterns and localized specular highlights, that are not resolved in imagery acquired from 20 m AGL. Reconstructing these features across a large altitude gap thus increases the risk of plausible-looking but physically ungrounded texture hallucination and artifacts, especially for HR areas without counterparts in LR. Since the adversarial loss was applied exclusively to the 10 m target, adversarial supervision may further encourage the synthesis of high-frequency textures, making the trade-off between perceptual sharpness and reconstruction fidelity particularly pronounced in this group. The elevated NIQE scores reported in Table \ref{tab:real_model_real} are consistent with the observed deterioration in visual quality, although they do not directly quantify hallucination. At moderate scale factors such as 70$\rightarrow$30 m of 2.37$\times$ scaling and 50$\rightarrow$20 m of 2.55$\times$ scaling, the adversarial loss was not applied to these pairs, since the target resolution does not demand fine-grained details, such as sub-leaf details. Under these settings, outputs are noticeably smoother but exhibit substantially less hallucination, with canopy row structure and inter-plant spacing well preserved.

\begin{table*}[t]
\centering
\setlength{\tabcolsep}{2pt}
\caption{SR evaluation of the proposed model trained on real-world data and evaluated on the real-world test set, grouped by target altitude. \textbf{Bold}: best value within each group. MAE is computed on intensities normalized to $[0,1]$, and Mean is computed over all 21 pairs. The lower panel reports the no-reference NIQE, which was computed only for the 10\,m target and is therefore averaged over 6 pairs.}
\label{tab:real_model_real}
\begin{tabular*}{\linewidth}{@{\extracolsep{\fill}}lll*{6}{c}@{}}
\toprule
\textbf{Target} & \textbf{Source} & \textbf{Scale} & \textbf{PSNR} (dB)\,$\uparrow$ & \textbf{SSIM}\,$\uparrow$ & \textbf{MS-SSIM}\,$\uparrow$ & \textbf{MAE}\,$\downarrow$ & \textbf{LPIPS}\,$\downarrow$ & \textbf{DISTS}\,$\downarrow$ \\
\midrule
\multirow{6}{*}{10\,m} & 20\,m & 1.67$\times$ & \textbf{16.04} & \textbf{0.2684} & \textbf{0.6497} & \textbf{0.1263} & \textbf{0.2236} & \textbf{0.2431} \\
 & 30\,m & 2.55$\times$ & 15.62 & 0.2134 & 0.6012 & 0.1337 & 0.2864 & 0.2698 \\
 & 40\,m & 3.42$\times$ & 15.63 & 0.1978 & 0.5834 & 0.1337 & 0.3311 & 0.2863 \\
 & 50\,m & 4.24$\times$ & 15.64 & 0.1846 & 0.5635 & 0.1344 & 0.3485 & 0.3002 \\
 & 60\,m & 5.09$\times$ & 15.55 & 0.1810 & 0.5534 & 0.1365 & 0.3910 & 0.3114 \\
 & 70\,m & 6.03$\times$ & 15.56 & 0.1751 & 0.5415 & 0.1366 & 0.4315 & 0.3291 \\
\midrule
\multirow{5}{*}{20\,m} & 30\,m & 1.53$\times$ & \textbf{19.53} & \textbf{0.5239} & \textbf{0.8170} & \textbf{0.0852} & \textbf{0.0989} & \textbf{0.1905} \\
 & 40\,m & 2.05$\times$ & 18.08 & 0.3933 & 0.7453 & 0.1013 & 0.1393 & 0.2418 \\
 & 50\,m & 2.55$\times$ & 17.56 & 0.3257 & 0.7041 & 0.1077 & 0.1645 & 0.2607 \\
 & 60\,m & 3.05$\times$ & 17.41 & 0.3229 & 0.6945 & 0.1103 & 0.1828 & 0.2651 \\
 & 70\,m & 3.62$\times$ & 17.24 & 0.2898 & 0.6706 & 0.1123 & 0.1967 & 0.2791 \\
\midrule
\multirow{4}{*}{30\,m} & 40\,m & 1.35$\times$ & \textbf{18.95} & \textbf{0.5265} & \textbf{0.8094} & \textbf{0.0934} & \textbf{0.0906} & \textbf{0.2185} \\
 & 50\,m & 1.67$\times$ & 18.33 & 0.4565 & 0.7733 & 0.0999 & 0.1137 & 0.2414 \\
 & 60\,m & 2.00$\times$ & 18.23 & 0.4590 & 0.7672 & 0.1016 & 0.1271 & 0.2488 \\
 & 70\,m & 2.37$\times$ & 17.77 & 0.4032 & 0.7365 & 0.1065 & 0.1417 & 0.2645 \\
\midrule
\multirow{3}{*}{40\,m} & 50\,m & 1.24$\times$ & \textbf{19.54} & 0.5805 & \textbf{0.8589} & \textbf{0.0854} & \textbf{0.1000} & \textbf{0.2242} \\
 & 60\,m & 1.49$\times$ & 19.45 & \textbf{0.5822} & 0.8528 & 0.0872 & 0.1102 & 0.2346 \\
 & 70\,m & 1.76$\times$ & 18.87 & 0.5263 & 0.8313 & 0.0925 & 0.1228 & 0.2520 \\
\midrule
\multirow{2}{*}{50\,m} & 60\,m & 1.20$\times$ & \textbf{19.75} & \textbf{0.6214} & \textbf{0.8210} & \textbf{0.0843} & \textbf{0.1114} & \textbf{0.2157} \\
 & 70\,m & 1.42$\times$ & 19.46 & 0.6005 & 0.8082 & 0.0868 & 0.1203 & 0.2245 \\
\midrule
60\,m & 70\,m & 1.18$\times$ & \textbf{20.81} & \textbf{0.7310} & \textbf{0.8980} & \textbf{0.0740} & \textbf{0.0940} & \textbf{0.2024} \\
\midrule
\multicolumn{3}{l}{\textbf{Mean}} & 17.86 & 0.4078 & 0.7277 & 0.1062 & 0.1870 & 0.2526 \\
\bottomrule
\end{tabular*}
\vspace{0.9em}
\begin{tabular*}{\linewidth}{@{\extracolsep{\fill}}l*{7}{c}@{}}
\toprule
\textbf{NIQE}\,$\downarrow$ & \multicolumn{6}{c}{\textbf{Source}} & \multirow{2}{*}{\textbf{Mean}} \\
\cmidrule(lr){2-7}
(\textbf{Target}\,$=$\,10\,m) & \textbf{20\,m} & \textbf{30\,m} & \textbf{40\,m} & \textbf{50\,m} & \textbf{60\,m} & \textbf{70\,m} & \\
\midrule
\textbf{Real} & 8.9020 & \textbf{8.8033} & 9.2837 & 9.2596 & 9.0440 & 9.3105 & 9.1005 \\
\bottomrule
\end{tabular*}
\vspace{1mm}
\begin{minipage}{\textwidth}
\centering
\footnotesize
$\uparrow$ indicates that higher is better, and $\downarrow$ indicates that lower is better.
\end{minipage}

\end{table*}

\subsection{Ablation Analyses and Conditioning Pathway Analysis}
\noindent
This section evaluates the conditioning design through a GSD Fourier encoder ablation and inference-time conditioning pathway analysis. For the GSD Fourier encoder ablation, the model was retrained under the same training configuration with explicit source and target GSD conditioning removed. In contrast, FiLM and the content-aware encoder are built into the conditioning pathway used by the unified model to accommodate different source-target GSD pairs. Removing these components and retraining would require modification of the conditioning architecture to maintain a functional multi-GSD model, introducing additional design choices into the comparison. These components were therefore analyzed through inference-time interventions on the trained full model. Specifically, FiLM, content-aware encoder, or both were disabled while the remaining model computation was retained. Quantitative results on these three variants compared to the full model across all source-target GSD pairs are reported in Table \ref{tab:ablation_and_conditioning}, in which \textbf{w/o FiLM} removes the scale-conditioned FiLM modulation, \textbf{w/o CaE} has the content-aware encoder disabled, \textbf{w/o Both} removes both FiLM and content-aware encoder, and \textbf{w/o GSD} removes the GSD Fourier encoder, so that neither the scale-conditioned FiLM nor expert-head routing receives the GSD embeddings originally from the GSD Fourier encoder.

\begin{sidewaystable*}
\centering
\caption{Ablation study and conditioning pathway analysis across 21 source-target altitude pairs. Bold values indicate the best performance in each column within each metric block. \textbf{CaE} denotes the content-aware encoder; "\textbf{Both}" denotes the joint removal of FiLM and CaE; "\textbf{GSD}" denotes the GSD Fourier encoder.}
\label{tab:ablation_and_conditioning}
\begingroup
\renewcommand{\arraystretch}{1.10}
\resizebox{\linewidth}{!}{%
\begin{tabular}{l|l|cccccc|ccccc|cccc|ccc|cc|c|c}
\toprule[2pt]
\textbf{Metric} & \textbf{Variant} & 70$\to$60 & 70$\to$50 & 70$\to$40 & 70$\to$30 & 70$\to$20 & 70$\to$10 & 60$\to$50 & 60$\to$40 & 60$\to$30 & 60$\to$20 & 60$\to$10 & 50$\to$40 & 50$\to$30 & 50$\to$20 & 50$\to$10 & 40$\to$30 & 40$\to$20 & 40$\to$10 & 30$\to$20 & 30$\to$10 & 20$\to$10 & \textbf{Mean} \\
\midrule
\multirow{5}{*}{PSNR$\uparrow$} & \textbf{Full} & \textbf{20.81} & 19.46 & 18.87 & 17.77 & \textbf{17.24} & \textbf{15.56} & \textbf{19.75} & \textbf{19.45} & \textbf{18.23} & \textbf{17.41} & \textbf{15.55} & \textbf{19.54} & \textbf{18.33} & \textbf{17.56} & \textbf{15.64} & \textbf{18.95} & \textbf{18.08} & \textbf{15.63} & \textbf{19.53} & \textbf{15.62} & 16.04 & \textbf{17.86} \\
& \textbf{w/o FiLM} & \abldeg{54.255}{20.47} & \abldeg{19.148}{19.34} & \abldeg{11.170}{18.80} & \abldeg{9.574}{17.71} & \abldeg{41.489}{16.98} & \abldeg{51.063}{15.24} & \abldeg{36.703}{19.52} & \abldeg{23.937}{19.30} & \abldeg{12.766}{18.15} & \abldeg{27.127}{17.24} & \abldeg{41.489}{15.29} & \abldeg{51.063}{19.22} & \abldeg{17.552}{18.22} & \abldeg{19.148}{17.44} & \abldeg{25.533}{15.48} & \abldeg{3.192}{18.93} & \abldeg{12.766}{18.00} & \abldeg{20.745}{15.50} & \abldeg{1.596}{19.52} & \abldeg{3.192}{15.60} & \ablimp{1.596}{\textbf{16.05}} & 17.71 \\
& \textbf{w/o CaE} & \abldeg{17.552}{20.70} & \abldeg{15.959}{19.36} & \abldeg{11.170}{18.80} & \abldeg{9.574}{17.71} & \abldeg{14.363}{17.15} & \abldeg{9.574}{15.50} & \abldeg{17.552}{19.64} & \abldeg{15.959}{19.35} & \abldeg{17.552}{18.12} & \abldeg{12.766}{17.33} & \abldeg{6.382}{15.51} & \abldeg{20.745}{19.41} & \abldeg{23.937}{18.18} & \abldeg{19.148}{17.44} & \abldeg{11.170}{15.57} & \abldeg{4.788}{18.92} & \abldeg{9.574}{18.02} & \abldeg{6.382}{15.59} & \abldeg{15.959}{19.43} & \abldeg{3.192}{15.60} & \abldeg{7.978}{15.99} & 17.78 \\
& \textbf{w/o Both} & \abldeg{75.000}{20.34} & \abldeg{27.127}{19.29} & \abldeg{19.148}{18.75} & \abldeg{11.170}{17.70} & \abldeg{41.489}{16.98} & \abldeg{51.063}{15.24} & \abldeg{44.681}{19.47} & \abldeg{28.723}{19.27} & \abldeg{14.363}{18.14} & \abldeg{27.127}{17.24} & \abldeg{41.489}{15.29} & \abldeg{57.448}{19.18} & \abldeg{19.148}{18.21} & \abldeg{20.745}{17.43} & \abldeg{23.937}{15.49} & \abldeg{4.788}{18.92} & \abldeg{17.552}{17.97} & \abldeg{27.127}{15.46} & \abldeg{3.192}{19.51} & \abldeg{3.192}{15.60} & \ablimp{1.596}{\textbf{16.05}} & 17.69 \\
& \textbf{w/o GSD} & \abldeg{11.170}{20.74} & \ablimp{17.552}{\textbf{19.57}} & \ablimp{7.978}{\textbf{18.92}} & \ablimp{6.382}{\textbf{17.81}} & \textbf{17.24} & \textbf{15.56} & \abldeg{3.192}{19.73} & \textbf{19.45} & \textbf{18.23} & \abldeg{4.788}{17.38} & \abldeg{4.788}{15.52} & \abldeg{12.766}{19.46} & \abldeg{1.596}{18.32} & \abldeg{7.978}{17.51} & \abldeg{3.192}{15.62} & \textbf{18.95} & \abldeg{4.788}{18.05} & \abldeg{4.788}{15.60} & \abldeg{9.574}{19.47} & \abldeg{4.788}{15.59} & \abldeg{1.596}{16.03} & 17.84 \\
\midrule
\multirow{5}{*}{SSIM$\uparrow$} & \textbf{Full} & \textbf{0.7310} & 0.6005 & \textbf{0.5263} & 0.4032 & \textbf{0.2898} & \textbf{0.1751} & \textbf{0.6214} & 0.5822 & \textbf{0.4590} & 0.3229 & 0.1810 & \textbf{0.5805} & \textbf{0.4565} & \textbf{0.3257} & 0.1846 & \textbf{0.5265} & 0.3933 & 0.1978 & 0.5239 & 0.2134 & 0.2684 & \textbf{0.4078} \\
& \textbf{w/o FiLM} & \abldeg{52.188}{0.7159} & \abldeg{17.627}{0.5954} & \abldeg{16.589}{0.5215} & \abldeg{11.405}{0.3999} & \abldeg{17.280}{0.2848} & \abldeg{10.022}{0.1722} & \abldeg{44.241}{0.6086} & \abldeg{29.032}{0.5738} & \abldeg{12.096}{0.4555} & \ablimp{1.036}{\textbf{0.3232}} & \ablimp{5.184}{0.1825} & \abldeg{57.028}{0.5640} & \abldeg{24.195}{0.4495} & \abldeg{6.220}{0.3239} & \ablimp{3.802}{0.1857} & \abldeg{9.333}{0.5238} & \abldeg{0.691}{0.3931} & \ablimp{2.766}{0.1986} & \ablimp{23.503}{0.5307} & \ablimp{36.291}{0.2239} & \ablimp{13.134}{\textbf{0.2722}} & 0.4047 \\
& \textbf{w/o CaE} & \abldeg{12.096}{0.7275} & \abldeg{11.405}{0.5972} & \abldeg{15.553}{0.5218} & \abldeg{14.170}{0.3991} & \abldeg{13.479}{0.2859} & \abldeg{5.184}{0.1736} & \abldeg{23.503}{0.6146} & \abldeg{19.702}{0.5765} & \abldeg{19.010}{0.4535} & \abldeg{14.170}{0.3188} & \abldeg{5.184}{0.1795} & \abldeg{33.180}{0.5709} & \abldeg{33.872}{0.4467} & \abldeg{24.195}{0.3187} & \abldeg{6.567}{0.1827} & \abldeg{12.096}{0.5230} & \abldeg{15.898}{0.3887} & \abldeg{6.912}{0.1958} & \abldeg{10.713}{0.5208} & \abldeg{8.986}{0.2108} & \abldeg{8.986}{0.2658} & 0.4034 \\
& \textbf{w/o Both} & \abldeg{75.000}{0.7093} & \abldeg{28.341}{0.5923} & \abldeg{23.503}{0.5195} & \abldeg{12.443}{0.3996} & \abldeg{12.443}{0.2862} & \abldeg{4.840}{0.1737} & \abldeg{51.152}{0.6066} & \abldeg{31.106}{0.5732} & \abldeg{13.479}{0.4551} & \ablimp{1.036}{\textbf{0.3232}} & \ablimp{5.529}{\textbf{0.1826}} & \abldeg{62.557}{0.5624} & \abldeg{27.996}{0.4484} & \abldeg{7.950}{0.3234} & \ablimp{4.493}{\textbf{0.1859}} & \abldeg{11.752}{0.5231} & \ablimp{1.383}{\textbf{0.3937}} & \ablimp{7.259}{\textbf{0.1999}} & \ablimp{29.723}{\textbf{0.5325}} & \ablimp{49.425}{\textbf{0.2277}} & \ablimp{12.443}{0.2720} & 0.4043 \\
& \textbf{w/o GSD} & \abldeg{9.333}{0.7283} & \ablimp{15.206}{\textbf{0.6049}} & \abldeg{5.529}{0.5247} & \ablimp{0.691}{\textbf{0.4034}} & \abldeg{2.419}{0.2891} & \abldeg{1.727}{0.1746} & \abldeg{2.419}{0.6207} & \ablimp{3.110}{\textbf{0.5831}} & \abldeg{2.766}{0.4582} & \abldeg{0.345}{0.3228} & \ablimp{0.345}{0.1811} & \abldeg{19.355}{0.5749} & \abldeg{5.184}{0.4550} & \abldeg{15.206}{0.3213} & \abldeg{2.419}{0.1839} & \abldeg{11.060}{0.5233} & 0.3933 & \abldeg{2.074}{0.1972} & \abldeg{1.727}{0.5234} & \ablimp{3.802}{0.2145} & \ablimp{2.419}{0.2691} & 0.4070 \\
\midrule
\multirow{5}{*}{MS-SSIM$\uparrow$} & \textbf{Full} & \textbf{0.8980} & 0.8082 & 0.8313 & \textbf{0.7365} & \textbf{0.6706} & \textbf{0.5415} & \textbf{0.8210} & 0.8528 & \textbf{0.7672} & \textbf{0.6945} & \textbf{0.5534} & \textbf{0.8589} & \textbf{0.7733} & \textbf{0.7041} & \textbf{0.5635} & \textbf{0.8094} & \textbf{0.7453} & \textbf{0.5834} & 0.8170 & 0.6012 & 0.6497 & \textbf{0.7277} \\
& \textbf{w/o FiLM} & \abldeg{45.652}{0.8910} & \abldeg{17.609}{0.8055} & \abldeg{11.740}{0.8295} & \abldeg{24.131}{0.7328} & \abldeg{53.477}{0.6624} & \abldeg{69.783}{0.5308} & \abldeg{52.826}{0.8129} & \abldeg{30.652}{0.8481} & \abldeg{28.043}{0.7629} & \abldeg{25.434}{0.6906} & \abldeg{34.566}{0.5481} & \abldeg{57.391}{0.8501} & \abldeg{36.523}{0.7677} & \abldeg{28.695}{0.6997} & \abldeg{24.783}{0.5597} & \abldeg{10.434}{0.8078} & \abldeg{6.523}{0.7443} & \abldeg{18.260}{0.5806} & \ablimp{17.609}{0.8197} & \ablimp{20.217}{0.6043} & \ablimp{6.523}{\textbf{0.6507}} & 0.7238 \\
& \textbf{w/o CaE} & \abldeg{18.260}{0.8952} & \abldeg{17.609}{0.8055} & \abldeg{18.260}{0.8285} & \abldeg{24.131}{0.7328} & \abldeg{25.434}{0.6667} & \abldeg{14.348}{0.5393} & \abldeg{30.652}{0.8163} & \abldeg{28.695}{0.8484} & \abldeg{36.523}{0.7616} & \abldeg{33.914}{0.6893} & \abldeg{18.260}{0.5506} & \abldeg{35.869}{0.8534} & \abldeg{48.260}{0.7659} & \abldeg{43.695}{0.6974} & \abldeg{21.523}{0.5602} & \abldeg{12.391}{0.8075} & \abldeg{19.566}{0.7423} & \abldeg{10.434}{0.5818} & \abldeg{9.131}{0.8156} & \abldeg{8.477}{0.5999} & \abldeg{7.826}{0.6485} & 0.7241 \\
& \textbf{w/o Both} & \abldeg{75.000}{0.8865} & \abldeg{36.523}{0.8026} & \abldeg{24.131}{0.8276} & \abldeg{26.086}{0.7325} & \abldeg{42.391}{0.6641} & \abldeg{50.869}{0.5337} & \abldeg{65.217}{0.8110} & \abldeg{37.174}{0.8471} & \abldeg{32.609}{0.7622} & \abldeg{27.391}{0.6903} & \abldeg{35.217}{0.5480} & \abldeg{66.523}{0.8487} & \abldeg{41.740}{0.7669} & \abldeg{30.652}{0.6994} & \abldeg{22.174}{0.5601} & \abldeg{13.043}{0.8074} & \abldeg{3.260}{0.7448} & \abldeg{7.174}{0.5823} & \ablimp{22.826}{\textbf{0.8205}} & \ablimp{34.566}{\textbf{0.6065}} & \ablimp{5.869}{0.6506} & 0.7235 \\
& \textbf{w/o GSD} & \abldeg{7.826}{0.8968} & \ablimp{18.260}{\textbf{0.8110}} & \ablimp{2.609}{\textbf{0.8317}} & \abldeg{1.957}{0.7362} & \abldeg{9.131}{0.6692} & \abldeg{4.566}{0.5408} & \abldeg{3.260}{0.8205} & \ablimp{3.260}{\textbf{0.8533}} & \abldeg{1.305}{0.7670} & \textbf{0.6945} & \abldeg{0.652}{0.5533} & \abldeg{16.957}{0.8563} & \abldeg{5.217}{0.7725} & \abldeg{11.740}{0.7023} & \abldeg{5.217}{0.5627} & \abldeg{12.391}{0.8075} & \abldeg{2.609}{0.7449} & \abldeg{5.217}{0.5826} & \abldeg{3.260}{0.8165} & \ablimp{6.523}{0.6022} & \ablimp{3.914}{0.6503} & 0.7272 \\
\midrule
\multirow{5}{*}{MAE$\downarrow$} & \textbf{Full} & \textbf{0.0740} & 0.0868 & 0.0925 & 0.1065 & \textbf{0.1123} & \textbf{0.1366} & \textbf{0.0843} & \textbf{0.0872} & \textbf{0.1016} & \textbf{0.1103} & \textbf{0.1365} & \textbf{0.0854} & \textbf{0.0999} & \textbf{0.1077} & \textbf{0.1344} & \textbf{0.0934} & \textbf{0.1013} & \textbf{0.1337} & \textbf{0.0852} & 0.1337 & \textbf{0.1263} & \textbf{0.1062} \\
 & \textbf{w/o FiLM} & \abldeg{57.558}{0.0773} & \abldeg{22.674}{0.0881} & \abldeg{12.209}{0.0932} & \abldeg{6.977}{0.1069} & \abldeg{50.581}{0.1152} & \abldeg{55.814}{0.1398} & \abldeg{41.860}{0.0867} & \abldeg{26.163}{0.0887} & \abldeg{13.953}{0.1024} & \abldeg{33.140}{0.1122} & \abldeg{45.349}{0.1391} & \abldeg{59.302}{0.0888} & \abldeg{17.442}{0.1009} & \abldeg{17.442}{0.1087} & \abldeg{15.698}{0.1353} & \abldeg{3.488}{0.0936} & \abldeg{15.698}{0.1022} & \abldeg{22.674}{0.1350} & \textbf{0.0852} & \ablimp{1.744}{0.1336} & \abldeg{1.744}{0.1264} & 0.1076 \\
 & \textbf{w/o CaE} & \abldeg{15.698}{0.0749} & \abldeg{13.953}{0.0876} & \abldeg{10.465}{0.0931} & \abldeg{8.721}{0.1070} & \abldeg{17.442}{0.1133} & \abldeg{10.465}{0.1372} & \abldeg{17.442}{0.0853} & \abldeg{15.698}{0.0881} & \abldeg{19.186}{0.1027} & \abldeg{17.442}{0.1113} & \abldeg{3.488}{0.1367} & \abldeg{20.930}{0.0866} & \abldeg{24.419}{0.1013} & \abldeg{24.419}{0.1091} & \abldeg{13.953}{0.1352} & \abldeg{5.233}{0.0937} & \abldeg{15.698}{0.1022} & \abldeg{17.442}{0.1347} & \abldeg{15.698}{0.0861} & \abldeg{10.465}{0.1343} & \abldeg{24.419}{0.1277} & 0.1071 \\
 & \textbf{w/o Both} & \abldeg{75.000}{0.0783} & \abldeg{31.395}{0.0886} & \abldeg{20.930}{0.0937} & \abldeg{8.721}{0.1070} & \abldeg{50.581}{0.1152} & \abldeg{59.302}{0.1400} & \abldeg{50.581}{0.0872} & \abldeg{31.395}{0.0890} & \abldeg{15.698}{0.1025} & \abldeg{31.395}{0.1121} & \abldeg{45.349}{0.1391} & \abldeg{66.279}{0.0892} & \abldeg{19.186}{0.1010} & \abldeg{19.186}{0.1088} & \abldeg{15.698}{0.1353} & \abldeg{5.233}{0.0937} & \abldeg{20.930}{0.1025} & \abldeg{34.884}{0.1357} & \abldeg{1.744}{0.0853} & \ablimp{3.488}{\textbf{0.1335}} & \abldeg{3.488}{0.1265} & 0.1078 \\
 & \textbf{w/o GSD} & \abldeg{13.953}{0.0748} & \ablimp{15.698}{\textbf{0.0859}} & \ablimp{5.233}{\textbf{0.0922}} & \ablimp{6.977}{\textbf{0.1061}} & \abldeg{3.488}{0.1125} & \abldeg{1.744}{0.1367} & \abldeg{3.488}{0.0845} & \abldeg{1.744}{0.0873} & \abldeg{1.744}{0.1017} & \abldeg{8.721}{0.1108} & \abldeg{6.977}{0.1369} & \abldeg{17.442}{0.0864} & \abldeg{1.744}{0.1000} & \abldeg{12.209}{0.1084} & \abldeg{8.721}{0.1349} & \textbf{0.0934} & \abldeg{1.744}{0.1014} & \abldeg{10.465}{0.1343} & \abldeg{6.977}{0.0856} & \abldeg{8.721}{0.1342} & \abldeg{5.233}{0.1266} & 0.1064 \\
\midrule[1.5pt]
\multirow{5}{*}{LPIPS$\downarrow$} & \textbf{Full} & \textbf{0.0940} & \textbf{0.1203} & \textbf{0.1228} & \textbf{0.1417} & 0.1967 & 0.4315 & \textbf{0.1114} & \textbf{0.1102} & \textbf{0.1271} & 0.1828 & 0.3910 & \textbf{0.1000} & \textbf{0.1137} & 0.1645 & 0.3485 & \textbf{0.0906} & \textbf{0.1393} & 0.3311 & 0.0989 & 0.2864 & 0.2236 & \textbf{0.1870} \\
& \textbf{w/o FiLM} & \abldeg{11.531}{0.1049} & \abldeg{9.309}{0.1291} & \abldeg{3.914}{0.1265} & \abldeg{2.538}{0.1441} & \abldeg{8.145}{0.2044} & \abldeg{71.297}{0.4989} & \abldeg{11.002}{0.1218} & \abldeg{6.030}{0.1159} & \abldeg{2.749}{0.1297} & \abldeg{6.877}{0.1893} & \abldeg{73.943}{0.4609} & \abldeg{8.250}{0.1078} & \abldeg{2.011}{0.1156} & \ablimp{2.327}{\textbf{0.1623}} & \abldeg{62.941}{0.4080} & \abldeg{2.327}{0.0928} & \abldeg{0.316}{0.1396} & \abldeg{51.199}{0.3795} & \ablimp{1.798}{\textbf{0.0972}} & \abldeg{28.350}{0.3132} & \ablimp{0.316}{0.2233} & 0.2031 \\
& \textbf{w/o CaE} & \abldeg{3.490}{0.0973} & \abldeg{3.598}{0.1237} & \abldeg{3.384}{0.1260} & \abldeg{4.020}{0.1455} & \abldeg{2.433}{0.1990} & \ablimp{2.327}{\textbf{0.4293}} & \abldeg{2.749}{0.1140} & \abldeg{3.279}{0.1133} & \abldeg{3.703}{0.1306} & \abldeg{3.809}{0.1864} & \ablimp{2.222}{\textbf{0.3889}} & \abldeg{2.749}{0.1026} & \abldeg{4.020}{0.1175} & \abldeg{2.116}{0.1665} & \ablimp{2.644}{\textbf{0.3460}} & \abldeg{3.384}{0.0938} & \abldeg{3.173}{0.1423} & \ablimp{0.741}{\textbf{0.3304}} & \abldeg{2.116}{0.1009} & \abldeg{0.316}{0.2867} & \abldeg{2.962}{0.2264} & 0.1889 \\
& \textbf{w/o Both} & \abldeg{16.819}{0.1099} & \abldeg{12.588}{0.1322} & \abldeg{4.760}{0.1273} & \abldeg{2.962}{0.1445} & \abldeg{9.204}{0.2054} & \abldeg{72.567}{0.5001} & \abldeg{9.945}{0.1208} & \abldeg{5.606}{0.1155} & \abldeg{2.962}{0.1299} & \abldeg{7.298}{0.1897} & \abldeg{75.000}{0.4619} & \abldeg{8.674}{0.1082} & \abldeg{2.433}{0.1160} & \ablimp{2.222}{0.1624} & \abldeg{64.528}{0.4095} & \abldeg{3.068}{0.0935} & \abldeg{0.741}{0.1400} & \abldeg{54.267}{0.3824} & \ablimp{1.270}{0.0977} & \abldeg{45.274}{0.3292} & \ablimp{0.211}{0.2234} & 0.2047 \\
& \textbf{w/o GSD} & \abldeg{2.116}{0.0960} & \abldeg{0.635}{0.1209} & \abldeg{0.105}{0.1229} & \abldeg{1.798}{0.1434} & \ablimp{2.857}{\textbf{0.1940}} & \abldeg{11.742}{0.4426} & \abldeg{0.846}{0.1122} & \abldeg{1.692}{0.1118} & \abldeg{2.433}{0.1294} & \ablimp{0.316}{\textbf{0.1825}} & \abldeg{14.916}{0.4051} & \abldeg{2.857}{0.1027} & \abldeg{1.376}{0.1150} & \abldeg{1.692}{0.1661} & \abldeg{14.386}{0.3621} & \textbf{0.0906} & \abldeg{0.846}{0.1401} & \abldeg{8.780}{0.3394} & \abldeg{1.798}{0.1006} & \ablimp{1.903}{\textbf{0.2846}} & \ablimp{0.741}{\textbf{0.2229}} & 0.1898 \\
\midrule
\multirow{5}{*}{DISTS$\downarrow$} & \textbf{Full} & \textbf{0.2024} & 0.2245 & 0.2520 & 0.2645 & 0.2791 & \textbf{0.3291} & 0.2157 & 0.2346 & \textbf{0.2488} & \textbf{0.2651} & \textbf{0.3114} & \textbf{0.2242} & \textbf{0.2414} & 0.2607 & 0.3002 & \textbf{0.2185} & \textbf{0.2418} & \textbf{0.2863} & 0.1905 & 0.2698 & 0.2431 & \textbf{0.2526} \\
& \textbf{w/o FiLM} & \abldeg{28.308}{0.2121} & \abldeg{11.965}{0.2286} & \abldeg{7.296}{0.2545} & \abldeg{8.170}{0.2673} & \abldeg{15.466}{0.2844} & \abldeg{63.035}{0.3507} & \abldeg{15.176}{0.2209} & \abldeg{7.587}{0.2372} & \abldeg{5.545}{0.2507} & \abldeg{9.047}{0.2682} & \abldeg{71.498}{0.3359} & \abldeg{11.381}{0.2281} & \abldeg{2.334}{0.2422} & \ablimp{1.460}{0.2602} & \abldeg{66.830}{0.3231} & \abldeg{2.627}{0.2194} & \abldeg{4.085}{0.2432} & \abldeg{61.575}{0.3074} & \ablimp{9.338}{0.1873} & \abldeg{16.634}{0.2755} & \ablimp{3.502}{\textbf{0.2419}} & 0.2590 \\
& \textbf{w/o CaE} & \abldeg{7.003}{0.2048} & \abldeg{4.669}{0.2261} & \abldeg{7.003}{0.2544} & \abldeg{10.505}{0.2681} & \abldeg{11.674}{0.2831} & \abldeg{0.584}{0.3293} & \abldeg{5.252}{0.2175} & \abldeg{5.836}{0.2366} & \abldeg{7.296}{0.2513} & \abldeg{9.921}{0.2685} & \abldeg{1.751}{0.3120} & \abldeg{4.962}{0.2259} & \abldeg{6.712}{0.2437} & \abldeg{4.378}{0.2622} & \ablimp{0.293}{\textbf{0.3001}} & \abldeg{5.545}{0.2204} & \abldeg{7.296}{0.2443} & \abldeg{7.296}{0.2888} & \abldeg{13.132}{0.1950} & \abldeg{16.343}{0.2754} & \abldeg{16.343}{0.2487} & 0.2551 \\
& \textbf{w/o Both} & \abldeg{37.355}{0.2152} & \abldeg{16.343}{0.2301} & \abldeg{9.047}{0.2551} & \abldeg{9.338}{0.2677} & \abldeg{17.217}{0.2850} & \abldeg{63.326}{0.3508} & \abldeg{14.883}{0.2208} & \abldeg{6.420}{0.2368} & \abldeg{5.836}{0.2508} & \abldeg{11.381}{0.2690} & \abldeg{75.000}{0.3371} & \abldeg{12.841}{0.2286} & \abldeg{3.795}{0.2427} & \ablimp{1.460}{0.2602} & \abldeg{66.246}{0.3229} & \abldeg{3.795}{0.2198} & \abldeg{4.085}{0.2432} & \abldeg{59.824}{0.3068} & \ablimp{11.381}{\textbf{0.1866}} & \abldeg{27.724}{0.2793} & \ablimp{2.334}{0.2423} & 0.2596 \\
& \textbf{w/o GSD} & \abldeg{11.381}{0.2063} & \ablimp{0.877}{\textbf{0.2242}} & \ablimp{2.918}{\textbf{0.2510}} & \ablimp{1.167}{\textbf{0.2641}} & \ablimp{4.085}{\textbf{0.2777}} & \abldeg{0.877}{0.3294} & \ablimp{4.378}{\textbf{0.2142}} & \ablimp{0.877}{\textbf{0.2343}} & \abldeg{7.003}{0.2512} & \abldeg{0.293}{0.2652} & \abldeg{14.299}{0.3163} & \abldeg{10.214}{0.2277} & \abldeg{1.751}{0.2420} & \ablimp{2.334}{\textbf{0.2599}} & \abldeg{16.927}{0.3060} & \abldeg{3.795}{0.2198} & \abldeg{1.751}{0.2424} & \abldeg{7.880}{0.2890} & \abldeg{10.505}{0.1941} & \ablimp{6.712}{\textbf{0.2675}} & \ablimp{2.044}{0.2424} & 0.2536 \\
\bottomrule[2pt]
\end{tabular}%
}
\endgroup

\vspace{2mm}
{\scriptsize
\begin{tabular}{@{}p{33.67mm}p{33.67mm}p{33.67mm}@{}}
\raggedright Improvement
& \centering No change
& \raggedleft Degradation\tabularnewline
\multicolumn{3}{@{}c@{}}{%
\begin{tikzpicture}
\shade[
    left color=AblationOrange!75!white,
    middle color=white,
    right color=AblationBlue!75!white
] (0,0) rectangle (101mm,2.2mm);
\end{tikzpicture}%
}\\[-0.5mm]
\multicolumn{3}{c}{Color intensity indicates the magnitude of the difference relative to the full model.}
\end{tabular}\par}

\vspace{1mm}
\begin{minipage}{\linewidth}
\footnotesize
Shading represents differences relative to Full, with a symmetric color scale normalized separately for each metric by its maximum absolute difference across all altitude pairs and ablated variants. Full rows and Mean cells are unshaded. $\uparrow$: higher is better; $\downarrow$: lower is better.
\end{minipage}
\end{sidewaystable*}


Overall, the full model achieved the best average performance across all metrics. Performance differences were more pronounced for perceptual metrics than for pixel-based and structure-based metrics. PSNR and SSIM remained low and varied only marginally across all configurations, consistent with the residual spatial misalignment in the real-world LR-HR pairs and the sensitivity of these metrics to such misalignment. Among the conditioning pathways, FiLM had the largest individual effect. Disabling FiLM reduced PSNR on 20 of the 21 pairs, whereas the perceptual degradation was mainly on pairs targeting 10 m. These results indicate that the trained model substantially relies on FiLM-mediated feature modulation to support pair-specific reconstruction across different source-target GSD pairs. Although \textbf{w/o CaE} alone led to only modest performance degradation, \textbf{w/o Both}, which disabled both content-aware encoder and FiLM, exhibited the largest overall degradation. This suggests that the content-aware encoder provides complementary conditioning information within the learned feature conditioning pathway, although its independent effect is mild.

The \textbf{w/o GSD} ablation yielded modest reductions in average performance across all six metrics relative to the full model, although improvements were observed for some individual altitude pairs. Mean LPIPS increased from 0.1870 to 0.1898, and mean DISTS increased from 0.2526 to 0.2536. These results suggest that explicit GSD Fourier encoding contributes to overall reconstruction quality, although its benefits vary across source-target altitude pairs. For example, removing GSD Fourier encoding resulted in the largest absolute LPIPS increase of 0.0141 for the 60$\rightarrow$10 m pair and the largest absolute DISTS increase of 0.0058 for the 50$\rightarrow$10 m pair.

To further assess whether the expert heads exhibited meaningful specialization, an additional set of experiments was conducted, in which each pair's correct GSD expert was replaced with the most distant available expert. The resulting performance changes are reported in Table \ref{tab:mismatch}. Wrong routing degraded almost all six metrics on every pair, with a mean LPIPS increase of 0.0537. Degradation was observed across all altitude pairs and metrics, with sensitivity to misrouting varying across pairs, except for a marginal PSNR improvement of 0.01 dB for the 70$\rightarrow$30 m pair. Despite this PSNR improvement, the LPIPS for the 70$\rightarrow$30 m pair increased by 0.0662, indicating that fidelity and perceptual metrics did not respond uniformly to expert misrouting. 

The 30$\rightarrow$20 m pair exhibited the largest degradation in PSNR, SSIM, MS-SSIM, MAE, and DISTS, together with an LPIPS increase of 0.1021. The largest LPIPS increase of 0.1283 occurred for the 70$\rightarrow$10 m pair, indicating that replacing the 70 m source-GSD expert with the 20 m expert substantially impaired perceptual reconstruction quality for this pair. Together, these results suggest that the expert heads learn source-GSD-specific transformations, supporting the importance of matching expert selection to the source GSD for accurate reconstruction across altitude pairs. Table \ref{tab:mismatch_mean} in the Supplementary Material reports performance changes averaged over all five incorrect expert assignments. All six metrics deteriorated on average, with mean LPIPS and DISTS increases of 0.0241 and 0.0117, respectively, indicating that the adverse effects of misrouting were not limited to the most distant expert assignment.

\begin{table*}[!t]
\centering
\caption{Impact of wrong expert routing. For each pair, the correct GSD expert is replaced by the most distant available expert. $\Delta = \mathrm{mismatched} - \mathrm{correct~routing}$. Negative values indicate degradation for PSNR, SSIM, and MS-SSIM, whereas positive values indicate degradation for MAE, LPIPS, and DISTS. Orange indicates improvement and blue indicates degradation, accounting for each metric's direction.}
\label{tab:mismatch}
\begingroup
\renewcommand{\arraystretch}{1.10}
\resizebox{\textwidth}{!}{%
\begin{tabular}{ll|rr|rr|rr|rr|rr|rr}
\toprule[2pt]
Src & Tgt & \multicolumn{2}{c|}{PSNR (dB)$\uparrow$} & \multicolumn{2}{c|}{SSIM$\uparrow$} & \multicolumn{2}{c|}{MS-SSIM$\uparrow$} & \multicolumn{2}{c|}{MAE$\downarrow$} & \multicolumn{2}{c|}{LPIPS$\downarrow$} & \multicolumn{2}{c}{DISTS$\downarrow$} \\
 & & Correct & $\Delta$ & Correct & $\Delta$ & Correct & $\Delta$ & Correct & $\Delta$ & Correct & $\Delta$ & Correct & $\Delta$ \\
\midrule
\multirow{6}{*}{70\,m} & 60\,m & 20.81 & \abldeg{22.241}{$-0.43$} & 0.7310 & \abldeg{12.357}{$-0.0144$} & 0.8980 & \abldeg{14.894}{$-0.0084$} & 0.0740 & \abldeg{21.711}{$+0.0044$} & 0.0940 & \abldeg{7.716}{$+0.0132$} & 0.2024 & \abldeg{15.870}{$+0.0135$} \\
 & 50\,m & 19.46 & \abldeg{9.828}{$-0.19$} & 0.6005 & \abldeg{11.585}{$-0.0135$} & 0.8082 & \abldeg{16.489}{$-0.0093$} & 0.0868 & \abldeg{11.349}{$+0.0023$} & 0.1203 & \abldeg{18.005}{$+0.0308$} & 0.2245 & \abldeg{8.229}{$+0.0070$} \\
 & 40\,m & 18.87 & \abldeg{5.172}{$-0.10$} & 0.5263 & \abldeg{15.017}{$-0.0175$} & 0.8313 & \abldeg{16.489}{$-0.0093$} & 0.0925 & \abldeg{5.921}{$+0.0012$} & 0.1228 & \abldeg{49.454}{$+0.0846$} & 0.2520 & \abldeg{9.875}{$+0.0084$} \\
 & 30\,m & 17.77 & \ablimp{0.517}{$+0.01$} & 0.4032 & \abldeg{7.895}{$-0.0092$} & 0.7365 & \abldeg{16.844}{$-0.0095$} & 0.1065 & \abldeg{0.987}{$+0.0002$} & 0.1417 & \abldeg{38.698}{$+0.0662$} & 0.2645 & \abldeg{17.516}{$+0.0149$} \\
 & 20\,m & 17.24 & \abldeg{1.552}{$-0.03$} & 0.2898 & \abldeg{3.175}{$-0.0037$} & 0.6706 & \abldeg{14.184}{$-0.0080$} & 0.1123 & \abldeg{2.467}{$+0.0005$} & 0.1967 & \abldeg{38.640}{$+0.0661$} & 0.2791 & \abldeg{23.864}{$+0.0203$} \\
 & 10\,m & 15.56 & \abldeg{15.517}{$-0.30$} & 0.1751 & \abldeg{5.921}{$-0.0069$} & 0.5415 & \abldeg{20.035}{$-0.0113$} & 0.1366 & \abldeg{27.138}{$+0.0055$} & 0.4315 & \abldeg{75.000}{$+0.1283$} & 0.3291 & \abldeg{33.268}{$+0.0283$} \\
\midrule
\multirow{5}{*}{60\,m} & 50\,m & 19.75 & \abldeg{25.862}{$-0.50$} & 0.6214 & \abldeg{22.912}{$-0.0267$} & 0.8210 & \abldeg{29.078}{$-0.0164$} & 0.0843 & \abldeg{27.138}{$+0.0055$} & 0.1114 & \abldeg{6.898}{$+0.0118$} & 0.2157 & \abldeg{8.111}{$+0.0069$} \\
 & 40\,m & 19.45 & \abldeg{19.655}{$-0.38$} & 0.5822 & \abldeg{25.744}{$-0.0300$} & 0.8528 & \abldeg{23.404}{$-0.0132$} & 0.0872 & \abldeg{19.243}{$+0.0039$} & 0.1102 & \abldeg{29.754}{$+0.0509$} & 0.2346 & \abldeg{8.817}{$+0.0075$} \\
 & 30\,m & 18.23 & \abldeg{14.483}{$-0.28$} & 0.4590 & \abldeg{19.908}{$-0.0232$} & 0.7672 & \abldeg{34.397}{$-0.0194$} & 0.1016 & \abldeg{14.803}{$+0.0030$} & 0.1271 & \abldeg{32.736}{$+0.0560$} & 0.2488 & \abldeg{18.574}{$+0.0158$} \\
 & 20\,m & 17.41 & \abldeg{14.483}{$-0.28$} & 0.3229 & \abldeg{13.301}{$-0.0155$} & 0.6945 & \abldeg{28.014}{$-0.0158$} & 0.1103 & \abldeg{16.283}{$+0.0033$} & 0.1828 & \abldeg{20.109}{$+0.0344$} & 0.2651 & \abldeg{21.748}{$+0.0185$} \\
 & 10\,m & 15.55 & \abldeg{12.414}{$-0.24$} & 0.1810 & \abldeg{5.835}{$-0.0068$} & 0.5534 & \abldeg{19.326}{$-0.0109$} & 0.1365 & \abldeg{22.204}{$+0.0045$} & 0.3910 & \abldeg{59.626}{$+0.1020$} & 0.3114 & \abldeg{38.323}{$+0.0326$} \\
\midrule
\multirow{4}{*}{50\,m} & 40\,m & 19.54 & \abldeg{27.414}{$-0.53$} & 0.5805 & \abldeg{31.322}{$-0.0365$} & 0.8589 & \abldeg{34.397}{$-0.0194$} & 0.0854 & \abldeg{28.125}{$+0.0057$} & 0.1000 & \abldeg{12.101}{$+0.0207$} & 0.2242 & \abldeg{15.635}{$+0.0133$} \\
 & 30\,m & 18.33 & \abldeg{23.276}{$-0.45$} & 0.4565 & \abldeg{28.318}{$-0.0330$} & 0.7733 & \abldeg{37.234}{$-0.0210$} & 0.0999 & \abldeg{24.671}{$+0.0050$} & 0.1137 & \abldeg{16.426}{$+0.0281$} & 0.2414 & \abldeg{18.339}{$+0.0156$} \\
 & 20\,m & 17.56 & \abldeg{25.345}{$-0.49$} & 0.3257 & \abldeg{26.173}{$-0.0305$} & 0.7041 & \abldeg{39.362}{$-0.0222$} & 0.1077 & \abldeg{31.086}{$+0.0063$} & 0.1645 & \abldeg{11.282}{$+0.0193$} & 0.2607 & \abldeg{10.933}{$+0.0093$} \\
 & 10\,m & 15.64 & \abldeg{7.759}{$-0.15$} & 0.1846 & \abldeg{4.548}{$-0.0053$} & 0.5635 & \abldeg{13.121}{$-0.0074$} & 0.1344 & \abldeg{14.309}{$+0.0029$} & 0.3485 & \abldeg{50.273}{$+0.0860$} & 0.3002 & \abldeg{35.149}{$+0.0299$} \\
\midrule
\multirow{3}{*}{40\,m} & 30\,m & 18.95 & \abldeg{24.310}{$-0.47$} & 0.5265 & \abldeg{29.348}{$-0.0342$} & 0.8094 & \abldeg{35.461}{$-0.0200$} & 0.0934 & \abldeg{21.711}{$+0.0044$} & 0.0906 & \abldeg{20.226}{$+0.0346$} & 0.2185 & \abldeg{30.564}{$+0.0260$} \\
 & 20\,m & 18.08 & \abldeg{38.276}{$-0.74$} & 0.3933 & \abldeg{38.873}{$-0.0453$} & 0.7453 & \abldeg{41.489}{$-0.0234$} & 0.1013 & \abldeg{42.434}{$+0.0086$} & 0.1393 & \abldeg{27.475}{$+0.0470$} & 0.2418 & \abldeg{24.216}{$+0.0206$} \\
 & 10\,m & 15.63 & \abldeg{10.862}{$-0.21$} & 0.1978 & \abldeg{10.898}{$-0.0127$} & 0.5834 & \abldeg{21.454}{$-0.0121$} & 0.1337 & \abldeg{13.322}{$+0.0027$} & 0.3311 & \abldeg{38.465}{$+0.0658$} & 0.2863 & \abldeg{34.091}{$+0.0290$} \\
\midrule
\multirow{2}{*}{30\,m} & 20\,m & 19.53 & \abldeg{75.000}{$-1.45$} & 0.5239 & \abldeg{75.000}{$-0.0874$} & 0.8170 & \abldeg{75.000}{$-0.0423$} & 0.0852 & \abldeg{75.000}{$+0.0152$} & 0.0989 & \abldeg{59.684}{$+0.1021$} & 0.1905 & \abldeg{75.000}{$+0.0638$} \\
 & 10\,m & 15.62 & \abldeg{20.172}{$-0.39$} & 0.2134 & \abldeg{18.278}{$-0.0213$} & 0.6012 & \abldeg{32.979}{$-0.0186$} & 0.1337 & \abldeg{27.138}{$+0.0055$} & 0.2864 & \abldeg{21.863}{$+0.0374$} & 0.2698 & \abldeg{26.920}{$+0.0229$} \\
\midrule
\multirow{1}{*}{20\,m} & 10\,m & 16.04 & \abldeg{34.138}{$-0.66$} & 0.2684 & \abldeg{33.810}{$-0.0394$} & 0.6497 & \abldeg{44.149}{$-0.0249$} & 0.1263 & \abldeg{46.875}{$+0.0095$} & 0.2236 & \abldeg{25.078}{$+0.0429$} & 0.2431 & \abldeg{29.859}{$+0.0254$} \\
\midrule
\multicolumn{2}{l|}{\textbf{Mean $\Delta$}} & & $-0.39$ & & $-0.0244$ & & $-0.0163$ & & $+0.0048$ & & $+0.0537$ & & $+0.0205$ \\
\bottomrule[2pt]
\end{tabular}%
}
\endgroup
\vspace{2mm}

{\scriptsize
\begin{tabular}{@{}p{33.67mm}p{33.67mm}p{33.67mm}@{}}
\raggedright Improvement & \centering No change & \raggedleft Degradation\tabularnewline
\multicolumn{3}{@{}c@{}}{%
\begin{tikzpicture}
\shade[left color=AblationOrange!75!white,middle color=white,right color=AblationBlue!75!white]
(0,0) rectangle (101mm,2.2mm);
\end{tikzpicture}%
}\\[-0.5mm]
\multicolumn{3}{c}{Color intensity indicates the magnitude of the difference relative to the full model.}
\end{tabular}\par}
\vspace{1mm}

\begin{minipage}{\textwidth}
\footnotesize
Shading is applied only to $\Delta$ cells, with a symmetric color scale normalized separately for each metric by its maximum absolute $\Delta$ across all altitude pairs. Correct and Mean cells are unshaded. For PSNR, SSIM, and MS-SSIM, positive $\Delta$ indicates improvement; for MAE, LPIPS, and DISTS, negative $\Delta$ indicates improvement. $\uparrow$: higher is better; $\downarrow$: lower is better.
\end{minipage}
\end{table*}

\subsection{Feature Space Analysis}
\noindent
To examine whether the conditioned representation supplied to the source-GSD expert heads captures both image content and source GSD, the FiLM-modulated feature maps were visualized using t-distributed Stochastic Neighbor Embedding (t-SNE) \citep{van2008visualizing}. Each feature map was converted into an image-level embedding by global average pooling, reduced to 50 dimensions by principal component analysis, and projected into two dimensions by t-SNE. Since t-SNE preserves local neighborhoods rather than global distances, the analysis below focuses on which samples group together and their relative ordering rather than on inter-cluster distances.

\begin{figure*}[t!]
\centering
\includegraphics[width=0.75\textwidth]{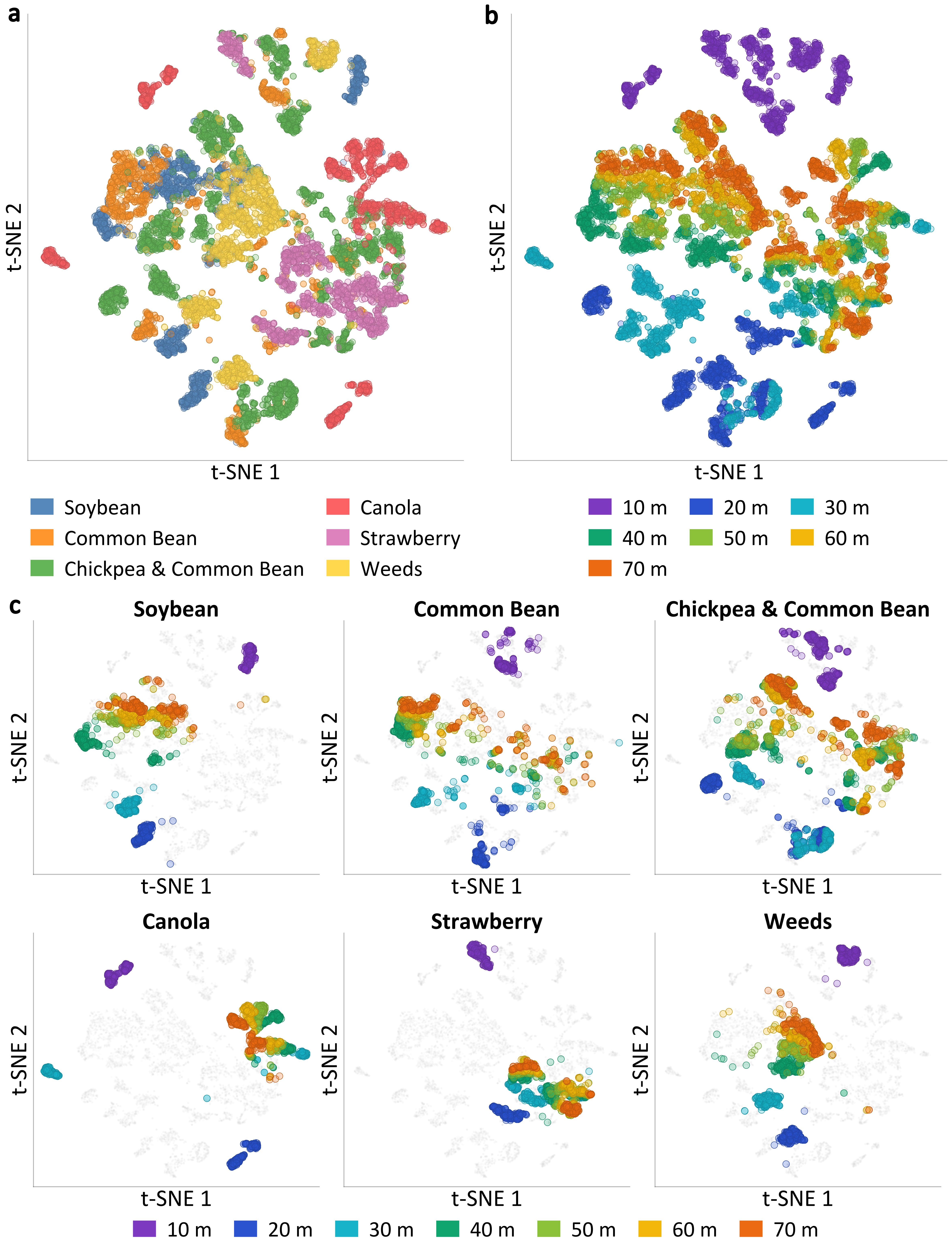} 
\caption{Conditioned feature space visualized by crop type and altitudes. (a) Points colored by crop types. (b) Points colored by input altitudes. (c) Per-crop points colored by input altitudes. }
\label{fig: tSNE}
\end{figure*}

Fig. \ref{fig: tSNE} presents three complementary views of the same t-SNE projection. In Fig. \ref{fig: tSNE}(a), the embeddings of different crop types form distinct clusters, even though no crop labels were used during training. These clusters indicate that the conditioned representation preserves scene-content variation associated with the crop categories in the dataset. When colored by source altitude as shown in Fig. \ref{fig: tSNE}(b), the projection exhibits an ordered transition from 10$\rightarrow$70 m, which is consistent with the continuous variation in GSD across acquisition altitudes. Images from 10 m AGL, which retain the finest spatial details, show the clearest separation from all other altitudes, whereas those acquired at 40 m to 70 m lie closer together as fine-scale information diminishes. 

Fig. \ref{fig: tSNE}(c) isolates each crop category within the same projection, making the within-crop altitude organization more apparent. Within each crop category, the 10 m images remain clearly distinct from the other altitudes, while the higher-altitude images exhibit more gradual transitions. The recurrence of this pattern across crop-specific panels indicates that the altitude-related organization is not restricted to a single crop category. Overall, the visualization shows that the conditioned feature space retains scene-content structure while systematically organizing samples by source GSD.

\subsection{Patch-level local attribution map with GSD-based physical interpretation}
\noindent
To interpret the spatial dependencies of the multi-altitude SR model, this study developed a variant of Local Attribution Map (LAM) \citep{gu2021interpreting}, combined with centroid-based Effective Receptive Field (ERF) radius estimation. This analysis examines whether GSD conditioning produces an interpretable adaptation in the spatial extent of the LR information used by the model, rather than functioning merely as an additional input parameter. Attribution was computed using Integrated Gradients (IG) from a Gaussian-blurred baseline, with a local high-frequency objective defined over a fixed query region in the reconstructed SR image. The resulting attribution was accumulated on the native LR input grid. The radius around the attribution centroid that enclosed 50\% of the attribution mass, denoted ERF50, was converted from source pixels to ground units using the source GSD. This allowed the effective input support to be compared across altitudes on a physical scale.

\begin{figure*}[!t]
\centering
\includegraphics[width=\textwidth]{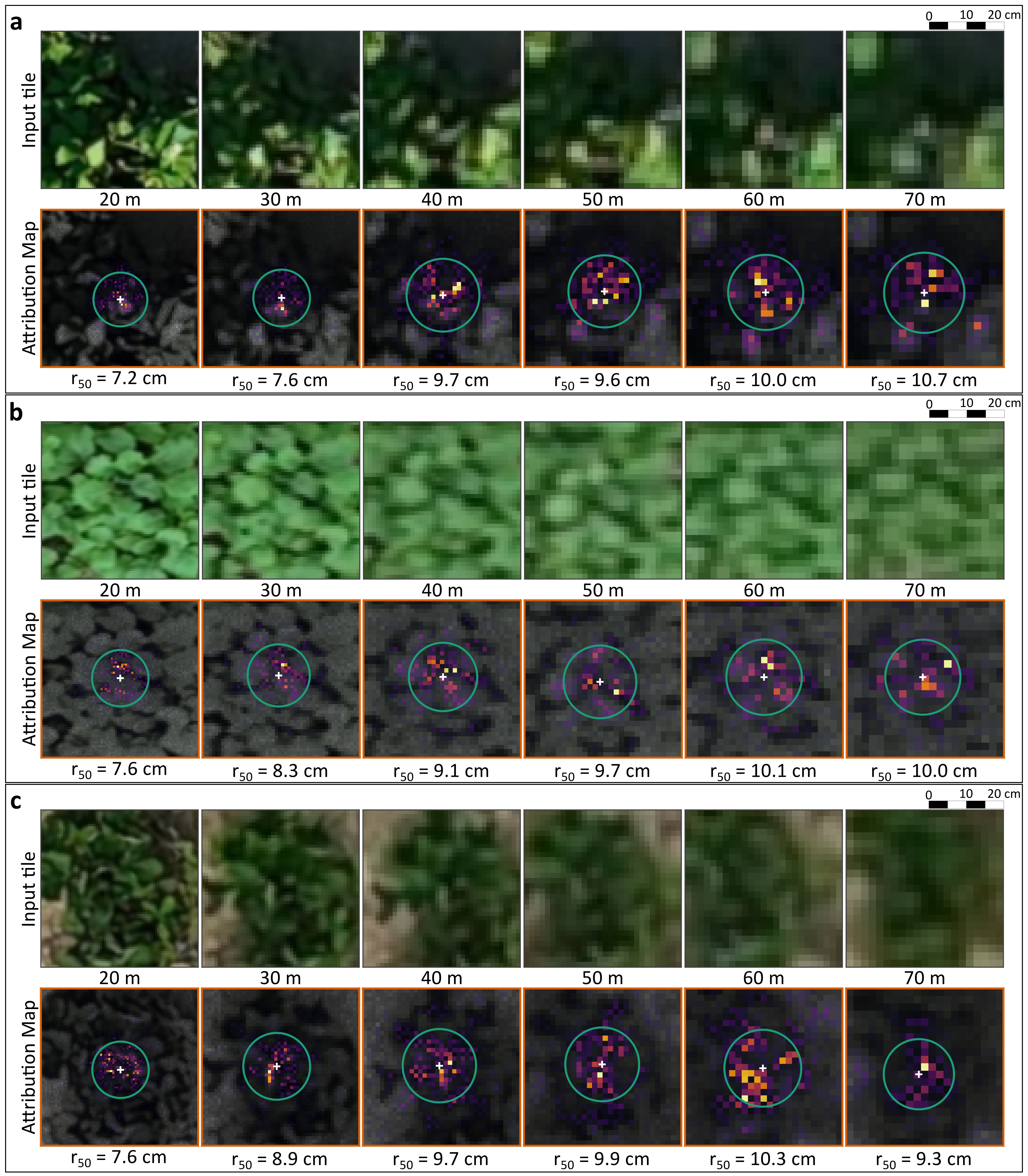} 
\caption{GSD-calibrated patch-level LAM of three crops across all source altitudes targeting the same 10 m HR. Panels (a), (b), and (c) are representative LAMs of common bean, canola, and strawberry, respectively. The white cross marks the attribution centroid, and the cyan circle is the $\text{ERF}_{50}$ radius that encloses 50\% of the attribution mass. }
\label{fig:LAM}
\end{figure*}

Fig.~\ref{fig:LAM} qualitatively illustrates this adaptation for three example scenes across source altitudes ranging from 20 to 70 m. At lower altitudes, the attribution is concentrated around a small number of localized image structures. As altitude increases and these structures become progressively less resolved, the attribution becomes spatially broader and less sharply peaked, indicating that the model integrates information over a larger portion of the available LR input. Despite this broader mass distribution, the attribution remains localized around the corresponding query region rather than becoming diffuse across the entire tile or shifting toward unrelated scene content. Quantitatively, the mean $\mathrm{ERF}_{50}$ radius decreases from $15.28\pm1.71$ pixels at 20~m to $5.38\pm0.64$ pixels at 70~m. In ground units, however, the corresponding footprint increases only from $8.41\pm0.94$ to $10.71\pm1.28$~cm, despite a $3.62\times$ increase in source GSD. Thus, the model uses fewer LR pixels at higher altitudes while maintaining a broadly similar physical support for reconstruction. These results suggest that altitude adaptation primarily involves changes in the extent of the input support measured in LR pixels, while the physical region used for reconstruction remains broadly consistent.

\subsection{Applications}
\begin{figure*}[!tp]
\centering
\includegraphics[width=0.8\textwidth]{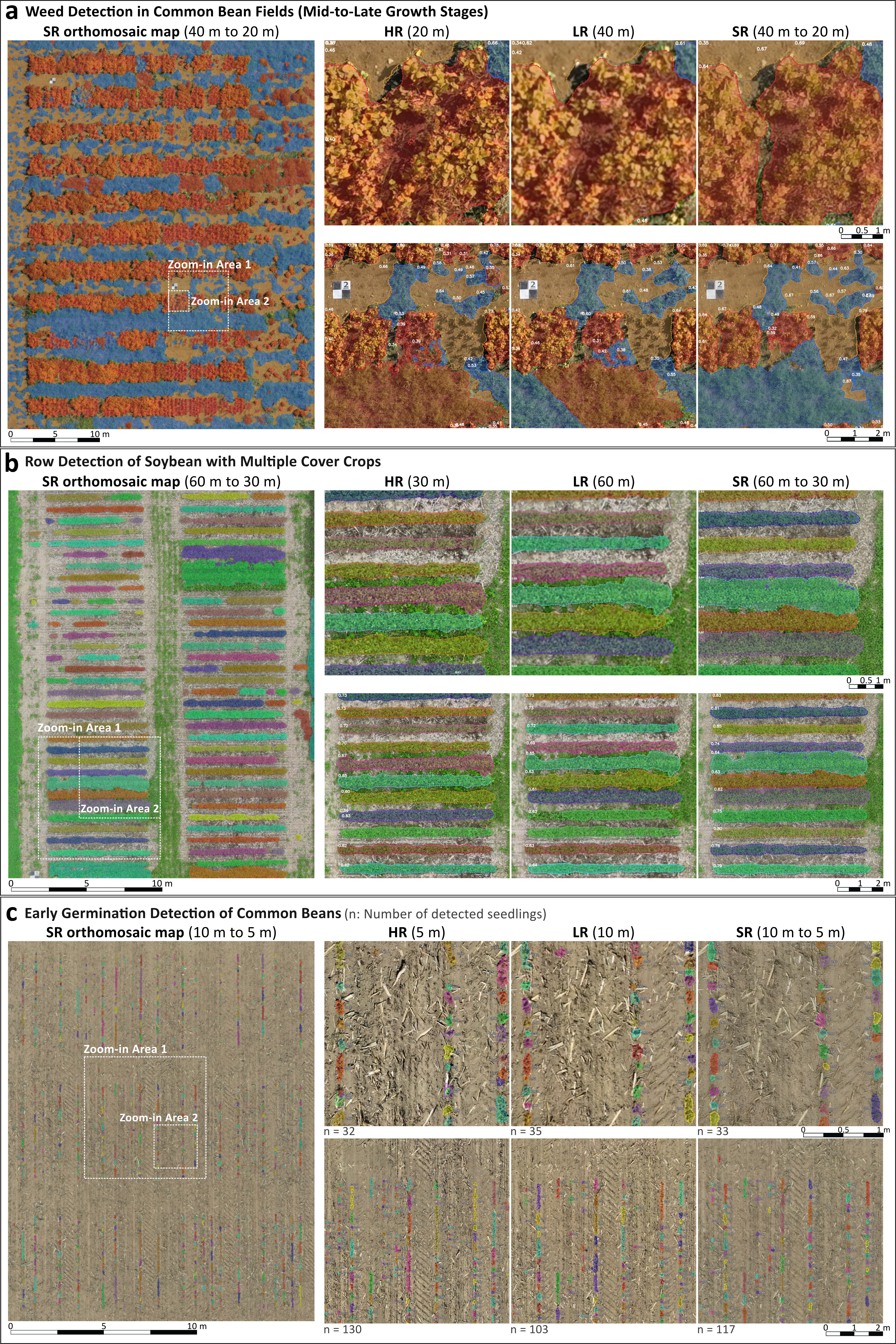} 
\caption{Zero-shot segmentation applications using SAM 3. (a) Weed detection in a common bean field at a middle-to-late growth stage, with SR from 40$\rightarrow$20 m. (b) Crop-row segmentation of soybean grown with multiple cover crops, with SR from 60$\rightarrow$30 m. (c) Detection of common bean seedlings at a very early growth stage, with SR from 10$\rightarrow$5 m. Numbered boxes mark the zoom-in areas discussed in the text.}
\label{fig: Applications}
\end{figure*}

\noindent
Fig. \ref{fig: Applications} presents three zero-shot segmentation applications using SAM 3 \citep{carion2025sam} on LR, SR, and HR imagery. These examples cover different crop arrangements, growth stages, and source-target altitude pairs, illustrating how the reconstructed spatial detail affects the detection and delineation of agricultural targets. Alongside a seedling detection example at fine resolution, the other two examples examine weed detection and crop-row segmentation at intermediate target altitudes of 20 and 30 m. 

Fig. \ref{fig: Applications}(a) shows weed detection in a common bean field at a middle-to-late growth stage using SR from 40$\rightarrow$20 m. Three text-prompt groups were used. Text prompts, "\textit{plant}", "\textit{crop}", "\textit{mature plant}", and "\textit{mature crop}", were used for common bean. The weeds were prompted as "\textit{grass}", and "\textit{soil}" for soil. Although SAM 3 struggled to identify common bean plants with dry or sparse foliage, most weeds were detected and distinguished from the crop. In zoom-in area 1, weeds in the lower part of the HR and LR images are incorrectly segmented as crops, but are correctly identified as weeds in SR. The segmentation on SR also resolves a gap between leaves in zoom-in area 2. These observations suggest that reconstruction to an intermediate resolution can help distinguish crops from weeds with contrasting visual characteristics. Changes in local texture or contrast introduced by SR may contribute to this effect.

Fig. \ref{fig: Applications}(b) shows a crop row detection application with soybean grown with multiple cover crops, none of which were represented in the SR training dataset. The text prompt "\textit{crop}" was used to segment soybean rows. Compared with LR, SR from 60$\rightarrow$30 m provides more complete segmentation of some relatively isolated soybean rows, including row portions omitted from the LR masks. However, improvements remain limited where soybean and cover crop canopies densely overlap. In these regions, SR does not properly separate the soybean rows from the surrounding vegetation, while HR provides clear segmentation of soybean rows, distinguishing them from the surrounding cover crops. This limitation may reflect both the loss of discriminative detail in LR and the absence of cover crops from the SR training dataset, potentially leading to reconstruction of broad vegetation patterns without sufficient detail to distinguish soybean rows from the surrounding cover crops.

Fig. \ref{fig: Applications}(c) shows common bean seedlings at a very early growth stage, which is also absent from the SR training dataset. Neither the source altitude of 10 m nor the target altitude of 5 m was included in the training pair. The input was routed to the nearest available source expert for 20 m. The text prompt "\textit{plant}" was used to detect individual seedlings. Despite this unseen combination of growth stage, source GSD, and target GSD, SR from 10$\rightarrow$5 m improves the visibility and detection of some seedlings. In zoom-in area 1, SAM 3 detects 117 instances in SR, closer to the 130 detected in HR than the 103 in LR. Zoom-in area 2 more clearly illustrates how segmentation on SR separates adjacent seedlings that are incorrectly merged into single instances in HR. For example, two adjacent seedlings in the bottom-right corner are correctly segmented as separate instances in SR but merged into a single instance in HR. This local improvement in instance separation may arise from changes in contrast or texture introduced by SR, making individual seedlings more distinguishable to the segmentation model. As with the weed example in Fig. \ref{fig: Applications}(a), these cases might reflect the sensitivity of SAM 3 to local contrast and texture rather than greater fidelity of SR relative to HR.

Together, these examples suggest that SR can support downstream segmentation under conditions both within and outside the current SR training distribution, with benefits that depend on the target and scene. The intermediate-resolution examples show that useful improvements do not necessarily require reconstruction to the finest available GSD. For large altitude gaps, such as 70$\rightarrow$10 m, the limited spatial information in the source imagery constrains reconstruction, and artifacts introduced by SR could adversely affect segmentation. By comparison, recovering sufficient row structure and boundary detail at intermediate resolutions may already support some agricultural tasks. The results for 40$\rightarrow$20 m and 60$\rightarrow$30 m suggest that the target GSD can be selected according to the application, while the remaining errors in densely overlapping vegetation highlight the limits of the detail recovered by SR.

\subsection{Discussion}
This study achieved GSD-aware multi-altitude SR with the proposed model, reconstructing 21 different GSD source-to-target pairs. For this study, the model is trained as a general-purpose model that is not especially optimized for any specific altitude pair. The 21 GSD pairs share the same RRDB feature-extraction backbone, with FiLM conditioning applied at two stages, in the content-aware encoder and on the backbone features. The SoTA comparison results show that the model achieves perceptual-based performance close to Real-ESRGAN, which is fine-tuned solely by the 20$\rightarrow$10 m pair. These results show that the model does not significantly harm the performance of a specific pair with the shared backbone. Thus, instead of 21 RRDB backbones each with over 16.5 M parameters, this study proposes a more computationally efficient structure that uses the GSD Fourier encoder and source-specific expert heads to distinguish these tasks.

This design also enables flexibility for future customization. For example, the current model has been trained with the multi-crop dataset and can be extended to additional crop types. Other unseen GSDs can also be integrated by adding lightweight experts for each new source GSD. While the current model assigns the same weight to all 21 pairs, a possible alternative is to assign primary and auxiliary weights to different pairs.

This study also looks into a less common usage of intermediate altitude reconstruction. In general computer vision, intermediate resolutions draw less attention compared to higher resolutions. However, this study shows that not only is targeting very high-resolution images like 10 m AGL meaningful, but also pairs such as 60$\rightarrow$30 m and 40$\rightarrow$20 m are of practical value in real-world applications. For large gap reconstructions, such as directly from 70 m to 10 m with 6.03$\times$ scaling, SR is likely to fail due to large scaling gaps.

Besides, most CV SR benchmarks focus on fixed integer scales such as 2$\times$, 3$\times$, and 4$\times$, paying less attention to the fractional scales that real UAV altitudes produce. In the proposed model, fractional scales are handled natively rather than rounding to the nearest integer factor. The model expects both target and source GSDs to be of continuous ranges, with the adaptive-upsampling kernels derived from $\text{log}(g_{src}/g_{tgt})$. The interpolation filter thus varies smoothly according to the fractional ratio. The proposed model achieves comparable perceptual quality compared to fixed-scale baselines for the 20$\rightarrow$10 m pairs, despite operating across 21 altitude pairs within a single unified framework. The lower SSIM than bicubic interpolation and lower PSNR than EDSR reflect the perception-distortion trade-off \citep{blau2018perception}. Since neither EDSR or bicubic attempts to reconstruct fine details, their smooth mean-regressed outputs result in higher fidelity metrics given the pixel misalignment in real multi-altitude pairs.

The quantitative results reveal a consistent pattern that reconstruction quality generally degrades as the GSD ratio between source and target altitude increases within each target group, with adjacent-altitude pairs almost always yielding the best performance. This behavior is expected, as larger altitude gaps entail a greater loss of information. For the 70→10 m pair, a single source pixel covers approximately 36 target pixels, so fine structures absent from the source images can only be inferred from learned priors rather than recovered.

The current framework supports multiple discrete source and target GSDs and can be easily extended by incorporating additional experts for more GSD conditions accordingly. Although multi-altitude SR is achieved through dual encoders, FiLM conditioning, and multiple GSD-aware reconstruction heads targeting each source GSD, the current model does not fully support continuous arbitrary-scale SR. Existing arbitrary-scale SR methods generally rely on implicit image representations, such as continuous image representation \citep{chen2021learning, lee2022local}, dynamic filter generation, and scale-aware upsampling mechanisms \citep{hu2019meta, wang2021learning}, which could be incorporated into the reconstruction stage in future work. Besides, our current model considers the inter-altitude and inter-reconstruction caused minor differences in GSD as negligible. For example, pixels in the same neighborhood might be of different GSDs due to the uneven surface on the ground. However, to achieve the next step of any-scale SR, these minor GSD differences might need to be taken into consideration but at the same time might cause heavier computational overhead.

Another limitation of the current model is the introduction of hallucinations and artifacts by its generative modules. In the current model, GAN is only used for pairs targeting 10 m AGL to create details and textures. Among these pairs, more pronounced artifacts were observed at larger scales. Based on our SoTA comparison, these artifacts generally appear in all models, especially with GAN integrated for fine-texture reconstruction. This study does not systematically investigate how to reduce these artifacts and how these artifacts will affect downstream agricultural tasks. This is left as a potential direction of investigation.

For our real-world agricultural UAV SR dataset, precise pixel-wise alignment is not applicable due to spatial and temporal changes when collecting multi-altitude data. However, this poses a known challenge in real production and deployment. This study shows these real-world degradation patterns are more complex than common simulated degradations, with all real sources of physical and reconstruction degradations. 

\section{Conclusion}
This study presents a GSD-conditioned SR framework for multi-altitude UAV imagery that addresses GSD-dependent degradation within a single unified model. To support this framework, a real-world multi-altitude crop dataset was constructed from UAV imagery captured at seven altitudes from 10 to 70 m AGL, providing 21 source-target altitude pairs with physically acquired rather than synthetically degraded LR inputs. On the 20$\rightarrow$10 m pair, the single proposed model achieves perceptual quality close to that of fixed-scale baselines fine-tuned exclusively for this pair, trailing Real-ESRGAN by only 4.3\% in LPIPS and 2.8\% in DISTS while achieving a better NIQE. Across all 21 pairs, reconstruction quality generally decreases as the source altitude increases within each target group, reflecting the growing information loss at larger GSD ratios. Ablation and conditioning pathway analyses show that FiLM modulation has the largest individual contribution, and that routing to the correct source-GSD expert is necessary for accurate reconstruction. Comparisons between synthetic and real-world LR imagery further reveal substantial discrepancies in their degradation characteristics, as a model trained on on-the-fly synthetic degradations yields lower PSNR, MS-SSIM, and higher MAE on all 21 real-world test pairs than the same model trained on real-world data. Qualitative downstream examples with zero-shot segmentation illustrate the practical value of selecting the target GSD according to the application, and an unseen 10$\rightarrow$5 m pair suggests preliminary potential for generalization beyond the training pairs. These results demonstrate that GSD-conditioned SR trained on real multi-altitude imagery is a practical approach for enhancing UAV imagery in precision agriculture. Future work will extend the framework toward continuous arbitrary-scale reconstruction and reduce the generative artifacts observed at large scale factors.

\section*{Acknowledgments}
This work was supported by the Fonds de recherche du Québec – Nature et technologies (FRQNT) and the Ministère de l’Agriculture, des Pêcheries et de l’Alimentation (MAPAQ) through the Research Partnership Program – Sustainable Agriculture (Grant No. 259806), the FRQNT and the Natural Sciences and Engineering Research Council of Canada (NSERC) through the NOVA–FRQNT–NSERC Program (Grant No. 269449), and the Institut de valorisation des données (IVADO) through its Exploratory Projects Program (Grant No. 269288). X.J. acknowledges support from NSERC through the Canada Graduate Research Scholarship – Doctoral (CGRS D).

The authors thank the Emile A. Lods Agronomy Research Centre at McGill University and Production Lareault Inc. for providing access to their fields and facilitating data collection.

\bibliographystyle{elsarticle-harv}
\bibliography{bibliography}

\clearpage
\onecolumn
\appendix
 \section*{Supplementary Material}
 \setcounter{figure}{0}
 \renewcommand{\thefigure}{S\arabic{figure}}
 \setcounter{table}{0}
 \renewcommand{\thetable}{S\arabic{table}}

\begin{table}[!htbp]
\centering
\caption{Impact of average-case expert misrouting. For each altitude pair, the reported $\Delta$ is the mean degradation averaged over all five incorrect expert assignments. $\Delta = \mathrm{mismatched} - \mathrm{correct routing}$. Negative values indicate degradation for PSNR, SSIM, and MS-SSIM, whereas positive values indicate degradation for MAE, LPIPS, and DISTS. Orange indicates improvement and blue indicates degradation, accounting for each metric's direction.}
\label{tab:mismatch_mean}
\begingroup
\renewcommand{\arraystretch}{1.10}
\resizebox{\textwidth}{!}{%
\begin{tabular}{ll|rr|rr|rr|rr|rr|rr}
\toprule[2pt]
Src & Tgt & \multicolumn{2}{c|}{PSNR (dB)$\uparrow$} & \multicolumn{2}{c|}{SSIM$\uparrow$} & \multicolumn{2}{c|}{MS-SSIM$\uparrow$} & \multicolumn{2}{c|}{MAE$\downarrow$} & \multicolumn{2}{c|}{LPIPS$\downarrow$} & \multicolumn{2}{c}{DISTS$\downarrow$} \\
 & & Correct & $\Delta$ & Correct & $\Delta$ & Correct & $\Delta$ & Correct & $\Delta$ & Correct & $\Delta$ & Correct & $\Delta$ \\
\midrule
\multirow{6}{*}{70\,m} & 60\,m & 20.81 & \abldeg{18.396}{$-0.26$} & 0.7310 & \abldeg{7.586}{$-0.0062$} & 0.8980 & \abldeg{9.596}{$-0.0038$} & 0.0740 & \abldeg{17.411}{$+0.0026$} & 0.0940 & \abldeg{8.402}{$+0.0082$} & 0.2024 & \abldeg{14.811}{$+0.0094$} \\
 & 50\,m & 19.46 & \abldeg{5.660}{$-0.08$} & 0.6005 & \abldeg{5.628}{$-0.0046$} & 0.8082 & \abldeg{9.848}{$-0.0039$} & 0.0868 & \abldeg{8.036}{$+0.0012$} & 0.1203 & \abldeg{11.680}{$+0.0114$} & 0.2245 & \abldeg{7.248}{$+0.0046$} \\
 & 40\,m & 18.87 & \ablimp{2.123}{$+0.03$} & 0.5263 & \abldeg{5.016}{$-0.0041$} & 0.8313 & \abldeg{8.081}{$-0.0032$} & 0.0925 & \ablimp{0.670}{$-0.0001$} & 0.1228 & \abldeg{27.357}{$+0.0267$} & 0.2520 & \abldeg{6.618}{$+0.0042$} \\
 & 30\,m & 17.77 & \ablimp{9.198}{$+0.13$} & 0.4032 & \ablimp{0.979}{$+0.0008$} & 0.7365 & \abldeg{5.556}{$-0.0022$} & 0.1065 & \ablimp{8.036}{$-0.0012$} & 0.1417 & \abldeg{27.766}{$+0.0271$} & 0.2645 & \abldeg{13.550}{$+0.0086$} \\
 & 20\,m & 17.24 & \ablimp{0.708}{$+0.01$} & 0.2898 & \ablimp{2.325}{$+0.0019$} & 0.6706 & \abldeg{7.828}{$-0.0031$} & 0.1123 & $+0.0000$ & 0.1967 & \abldeg{26.844}{$+0.0262$} & 0.2791 & \abldeg{11.345}{$+0.0072$} \\
 & 10\,m & 15.56 & \abldeg{9.906}{$-0.14$} & 0.1751 & \abldeg{2.936}{$-0.0024$} & 0.5415 & \abldeg{14.394}{$-0.0057$} & 0.1366 & \abldeg{16.071}{$+0.0024$} & 0.4315 & \abldeg{48.053}{$+0.0469$} & 0.3291 & \abldeg{13.235}{$+0.0084$} \\
\midrule
\multirow{5}{*}{60\,m} & 50\,m & 19.75 & \abldeg{29.717}{$-0.42$} & 0.6214 & \abldeg{25.693}{$-0.0210$} & 0.8210 & \abldeg{34.091}{$-0.0135$} & 0.0843 & \abldeg{33.482}{$+0.0050$} & 0.1114 & \abldeg{7.684}{$+0.0075$} & 0.2157 & \abldeg{5.357}{$+0.0034$} \\
 & 40\,m & 19.45 & \abldeg{19.104}{$-0.27$} & 0.5822 & \abldeg{23.369}{$-0.0191$} & 0.8528 & \abldeg{22.475}{$-0.0089$} & 0.0872 & \abldeg{18.750}{$+0.0028$} & 0.1102 & \abldeg{14.652}{$+0.0143$} & 0.2346 & \abldeg{7.878}{$+0.0050$} \\
 & 30\,m & 18.23 & \abldeg{7.783}{$-0.11$} & 0.4590 & \abldeg{17.985}{$-0.0147$} & 0.7672 & \abldeg{31.566}{$-0.0125$} & 0.1016 & \abldeg{11.384}{$+0.0017$} & 0.1271 & \abldeg{20.697}{$+0.0202$} & 0.2488 & \abldeg{15.599}{$+0.0099$} \\
 & 20\,m & 17.41 & \abldeg{7.783}{$-0.11$} & 0.3229 & \abldeg{9.421}{$-0.0077$} & 0.6945 & \abldeg{27.273}{$-0.0108$} & 0.1103 & \abldeg{11.384}{$+0.0017$} & 0.1828 & \abldeg{15.881}{$+0.0155$} & 0.2651 & \abldeg{14.023}{$+0.0089$} \\
 & 10\,m & 15.55 & \abldeg{9.906}{$-0.14$} & 0.1810 & \abldeg{3.915}{$-0.0032$} & 0.5534 & \abldeg{17.172}{$-0.0068$} & 0.1365 & \abldeg{16.071}{$+0.0024$} & 0.3910 & \abldeg{36.168}{$+0.0353$} & 0.3114 & \abldeg{27.416}{$+0.0174$} \\
\midrule
\multirow{4}{*}{50\,m} & 40\,m & 19.54 & \abldeg{29.009}{$-0.41$} & 0.5805 & \abldeg{34.258}{$-0.0280$} & 0.8589 & \abldeg{36.869}{$-0.0146$} & 0.0854 & \abldeg{30.804}{$+0.0046$} & 0.1000 & \abldeg{13.627}{$+0.0133$} & 0.2242 & \abldeg{10.872}{$+0.0069$} \\
 & 30\,m & 18.33 & \abldeg{13.443}{$-0.19$} & 0.4565 & \abldeg{22.757}{$-0.0186$} & 0.7733 & \abldeg{31.566}{$-0.0125$} & 0.0999 & \abldeg{16.071}{$+0.0024$} & 0.1137 & \abldeg{15.676}{$+0.0153$} & 0.2414 & \abldeg{14.023}{$+0.0089$} \\
 & 20\,m & 17.56 & \abldeg{12.736}{$-0.18$} & 0.3257 & \abldeg{17.618}{$-0.0144$} & 0.7041 & \abldeg{32.828}{$-0.0130$} & 0.1077 & \abldeg{17.411}{$+0.0026$} & 0.1645 & \abldeg{13.320}{$+0.0130$} & 0.2607 & \abldeg{9.454}{$+0.0060$} \\
 & 10\,m & 15.64 & \abldeg{6.368}{$-0.09$} & 0.1846 & \abldeg{2.936}{$-0.0024$} & 0.5635 & \abldeg{12.879}{$-0.0051$} & 0.1344 & \abldeg{10.714}{$+0.0016$} & 0.3485 & \abldeg{34.119}{$+0.0333$} & 0.3002 & \abldeg{23.792}{$+0.0151$} \\
\midrule
\multirow{3}{*}{40\,m} & 30\,m & 18.95 & \abldeg{18.396}{$-0.26$} & 0.5265 & \abldeg{25.449}{$-0.0208$} & 0.8094 & \abldeg{30.051}{$-0.0119$} & 0.0934 & \abldeg{18.080}{$+0.0027$} & 0.0906 & \abldeg{22.336}{$+0.0218$} & 0.2185 & \abldeg{23.950}{$+0.0152$} \\
 & 20\,m & 18.08 & \abldeg{28.302}{$-0.40$} & 0.3933 & \abldeg{30.832}{$-0.0252$} & 0.7453 & \abldeg{37.626}{$-0.0149$} & 0.1013 & \abldeg{31.473}{$+0.0047$} & 0.1393 & \abldeg{25.717}{$+0.0251$} & 0.2418 & \abldeg{18.120}{$+0.0115$} \\
 & 10\,m & 15.63 & \abldeg{7.783}{$-0.11$} & 0.1978 & \abldeg{7.586}{$-0.0062$} & 0.5834 & \abldeg{18.434}{$-0.0073$} & 0.1337 & \abldeg{10.714}{$+0.0016$} & 0.3311 & \abldeg{28.074}{$+0.0274$} & 0.2863 & \abldeg{28.361}{$+0.0180$} \\
\midrule
\multirow{2}{*}{30\,m} & 20\,m & 19.53 & \abldeg{75.000}{$-1.06$} & 0.5239 & \abldeg{75.000}{$-0.0613$} & 0.8170 & \abldeg{75.000}{$-0.0297$} & 0.0852 & \abldeg{75.000}{$+0.0112$} & 0.0989 & \abldeg{75.000}{$+0.0732$} & 0.1905 & \abldeg{75.000}{$+0.0476$} \\
 & 10\,m & 15.62 & \abldeg{11.321}{$-0.16$} & 0.2134 & \abldeg{10.889}{$-0.0089$} & 0.6012 & \abldeg{22.980}{$-0.0091$} & 0.1337 & \abldeg{15.402}{$+0.0023$} & 0.2864 & \abldeg{21.721}{$+0.0212$} & 0.2698 & \abldeg{22.216}{$+0.0141$} \\
\midrule
\multirow{1}{*}{20\,m} & 10\,m & 16.04 & \abldeg{26.179}{$-0.37$} & 0.2684 & \abldeg{25.204}{$-0.0206$} & 0.6497 & \abldeg{36.616}{$-0.0145$} & 0.1263 & \abldeg{36.830}{$+0.0055$} & 0.2236 & \abldeg{24.180}{$+0.0236$} & 0.2431 & \abldeg{25.053}{$+0.0159$} \\
\midrule
\multicolumn{2}{l|}{\textbf{Mean $\Delta$}} & & $-0.22$ & & $-0.0137$ & & $-0.0099$ & & $+0.0027$ & & $+0.0241$ & & $+0.0117$ \\
\bottomrule[2pt]
\end{tabular}%
}
\endgroup
\vspace{2mm}

{\scriptsize
\begin{tabular}{@{}p{33.67mm}p{33.67mm}p{33.67mm}@{}}
\raggedright Improvement & \centering No change & \raggedleft Degradation\tabularnewline
\multicolumn{3}{@{}c@{}}{%
\begin{tikzpicture}
\shade[left color=AblationOrange!75!white,middle color=white,right color=AblationBlue!75!white]
(0,0) rectangle (101mm,2.2mm);
\end{tikzpicture}%
}\\[-0.5mm]
\multicolumn{3}{c}{Color intensity indicates the magnitude of the difference relative to correct routing.}
\end{tabular}\par}
\vspace{1mm}

\begin{minipage}{\textwidth}
\footnotesize
Shading is applied only to $\Delta$ cells, with a symmetric color scale normalized separately for each metric by its maximum absolute $\Delta$ across all altitude pairs. Correct and Mean cells are unshaded. For PSNR, SSIM, and MS-SSIM, positive $\Delta$ indicates improvement; for MAE, LPIPS, and DISTS, negative $\Delta$ indicates improvement. $\uparrow$: higher is better; $\downarrow$: lower is better.
\end{minipage}
\end{table}

\end{document}